\documentclass{article}

\usepackage{eccvabbrv}
\usepackage{bm}

\usepackage{wrapfig}
\usepackage{graphicx}
\usepackage{booktabs}

\usepackage[accsupp]{axessibility}  

\usepackage{hyperref}
\usepackage{amssymb}
\usepackage{mathtools}
\usepackage{orcidlink}
\usepackage{multirow}
\usepackage{subcaption}
\usepackage{url}
\usepackage{authblk}

\usepackage[textsize=tiny]{todonotes}
\newtheorem{theorem}{Theorem}

\newtheorem{assumption}[theorem]{Assumption}
\newtheorem{definition}[theorem]{Definition}

\begin{document}

\title{Removing the Trigger, Not the Backdoor: Alternative Triggers and Latent Backdoors}

\author[1]{Gorka Abad}
\author[1]{Ermes Franch}
\author[2]{Stefanos Koffas}
\author[3,4]{Stjepan Picek}
\affil[1]{University of Bergen}
\affil[2]{Delft University of Technology}
\affil[3]{University of Zagreb}
\affil[4]{Radboud University}

\date{}

\maketitle

\begin{abstract}
    Current backdoor defenses assume that neutralizing a known trigger removes the backdoor. We show this trigger-centric view is incomplete: \emph{alternative triggers}, patterns perceptually distinct from training triggers, reliably activate the same backdoor. We estimate the alternative trigger backdoor direction in feature space by contrasting clean and triggered representations, and then develop a feature-guided attack that jointly optimizes target prediction and directional alignment. First, we theoretically prove that alternative triggers exist and are an inevitable consequence of backdoor training. Then, we verify this empirically. Additionally, defenses that remove training triggers often leave backdoors intact, and alternative triggers can exploit the latent backdoor feature-space. Our findings motivate defenses targeting backdoor directions in representation space rather than input-space triggers. 
\end{abstract}

\section{Introduction}

Backdoor attacks in neural networks (NNs) are among the most prominent security concerns~\cite{liu2020privacy}. By implanting a trigger into a subset of the training data, a backdoor is inserted into the model~\cite{gu2019badnets}. The backdoor then has two objectives: to perform as expected on clean data and to execute the malicious task when the trigger is present. Most existing backdoor defenses and evaluations focus on finding the trigger and then unlearning it~\cite{wang2019neural,liu2019abs,tao2022better}. Their view is trigger-centric: they assume that once the known trigger is neutralized, the backdoor is removed~\cite{li2021neural, wang2019neural, li2022backdoor}. 

Our starting point is that backdoors work by mapping multiple distinct pixel-space patterns into a shared malicious region in the model's feature space. This structure is analogous to a many-to-one mapping. Similar to a hash function that compresses a large input space into a fixed-size output space, a backdoored model compresses many distinct pixel-space patterns into a single malicious region in feature space. In the cryptographic setting, such many-to-one mappings are carefully designed to be collision-resistant, i.e., two different inputs should never produce the same output~\cite{bertoni2013keccak}. Backdoors have no such guarantee. This raises a natural question about backdoors: \textit{can backdoor triggers collide}?

This question is important for two reasons: (i) if triggers collide, defenders should not only defend against a single trigger. They must consider a wider range of possible triggers that will equally launch the malicious behavior. (ii) Attacks that create \textit{unique} triggers, often in the realm of sample-specific triggers~\cite{doan2021lira,li2021invisible}, typically define uniqueness as: the trigger is distinct in pixel space and tied to a specific sample or semantic pattern. However, our work suggests that uniqueness in pixel space does not imply uniqueness in feature space. The trigger is not truly unique from a defender's perspective, because alternative triggers can replicate the same malicious behavior.

In this paper, we provide novel insights into this problem. We demonstrate, from both theoretical and empirical perspectives, that alternative backdoor triggers exist. We support this claim with extensive experiments across multiple datasets (CIFAR-10/100 and TinyImagenet), models (ResNet-18 and VGG-19), attacks (BadNets~\cite{gu2019badnets}, Blend~\cite{chen2017targeted}, WaNet~\cite{nguyen2021wanet}, and Input-Aware~\cite{nguyen2020input}), and defenses (unlearning via fine-tuning, NAD~\cite{li2021neural}, and BAN~\cite{xu2024ban}). Our results show that even when defenses drop the attack success rate (ASR) of the original trigger to random guessing, alternative triggers still succeed with high ($>90\%$) ASR. This implies that the reverse-engineered trigger found by the defender is not enough to defend against the backdoor. 

Backdoors persist because poisoned training creates a persistent vulnerability in feature space, a backdoor region that remains accessible through multiple input patterns. We demonstrate this by estimating a backdoor direction via contrasting clean and triggered activations, and then developing a Feature-Guided Attack (FGA) that explicitly moves toward this direction while optimizing for the target class. We show that standard adversarial methods like Projected Gradient Descent (PGD), when targeted toward the backdoor label, could also discover alternative triggers with comparable success rates. By measuring the alignment between PGD-generated perturbations and our estimated backdoor direction, we find that both optimization strategies can converge toward the same feature-space region. FGA's contribution lies in providing a framework for verifying that discovered triggers genuinely exploit the backdoor mechanism, distinguishing backdoor-specific paths from arbitrary adversarial shortcuts.

Our contributions are: (i) a formalization showing that backdoor regions in feature space admit many alternative triggers, (ii) a feature-guided attack that systematically discovers alternative triggers, (iii) empirical evidence that alternative triggers remain effective after state-of-the-art defenses, and (iv) analysis showing that multiple optimization methods converge to the backdoor region.
\section{Related Work}

\paragraph{Backdoor attacks.}
Since the introduction of BadNets~\cite{gu2019badnets}, backdoor attacks have become a central research topic in adversarial machine learning. Subsequent work has explored various dimensions of the threat: stealthy triggers that are difficult to detect visually~\cite{nguyen2021wanet}, clean-label attacks where poisoned samples retain semantically consistent labels~\cite{turner2018clean}, and input-specific triggers that vary across samples~\cite{nguyen2020input}. Backdoor vulnerabilities have also been demonstrated across modalities and domains, including federated learning~\cite{bagdasaryan2020backdoor}, graph neural networks~\cite{zhang2021backdoor}, natural language processing~\cite{chen2021badnl}, and audio~\cite{koffas2022can}. Despite this diversity, existing work predominantly characterizes backdoors in terms of a specific trigger pattern and evaluates attacks by measuring the ASR of that pattern~\cite{xu2024ban,li2021neural,liu2018fine}.

\paragraph{Backdoor defenses.}
Defenses against backdoor attacks can be categorized by when they are executed, i.e., pre-training, in-training, and post-training, and by the domain in which they operate, i.e., input space, feature space, or weight space. Pre-training defenses~\cite{wang2019neural} detect and remove poisoned samples before model training, preventing backdoor implantation at the source. In-training defenses modify the learning algorithm or monitor internal activations to detect anomalies during optimization~\cite{wang2022trap}. Post-training defenses first detect whether a trained model is compromised~\cite{tran2018spectral} and then attempt to repair it via techniques such as pruning, fine-tuning, or activation suppression~\cite{li2021neural, xu2024ban}. While some defenses explicitly target feature-space representations~\cite{xu2024ban}, most evaluate success by measuring the ASR of the known trigger on a hold-out poisoned test set. This evaluation implicitly assumes that neutralizing the training trigger eliminates the backdoor.

\paragraph{Our contribution relative to prior work.}
Existing defenses are largely \emph{trigger-centric}: they reason about backdoors in terms of specific input patterns and consider a model safe once the ASR of the known trigger drops. We challenge this view by demonstrating that backdoors define persistent regions in feature space that can be activated by many distinct input patterns, i.e., alternative triggers, even after the original trigger has been successfully suppressed. While recent work has explored adversarial perturbations as probes for backdoor detection~\cite{xugrond}, we formalize the existence of alternative triggers and introduce a feature-guided attack that systematically discovers them by aligning with the backdoor direction in latent space. Critically, we show that state-of-the-art post-training defenses, including those that operate in feature space~\cite{xu2024ban}, do not fully erase this latent backdoor mechanism, leaving models vulnerable to feature-aligned attacks. Our work shifts the focus from trigger patterns to the underlying feature-space structure and highlights the need for defenses that explicitly target and erase the backdoor region itself.

\section{Backdoor Attacks on Neural Networks}

We define a neural network as a function $f_\theta(\cdot)$ parametrized by $\theta$ that takes an input $\bm x \in \mathcal{X} \in \mathbb{R}^n$ and produces an output $\bm y \in \mathcal{Y} \in \mathbb{R}^m$. Thus, a NN is a mapping $f(\cdot): \mathcal{X} \to \mathcal{Y}$ composed with concatenated layers $f(\cdot) \equiv f_{0} \circ f_{1} \circ f_{2} \circ \dots  \circ f_{l-1}$ where $l$ is layer number. Note that we omit $\theta$ to ease reading unless explicitly needed. The detailed notation is provided in Table~\ref{tab:notation} in Appendix~\ref{sec:notation}.




Backdoor attacks~\cite{gu2019badnets} are training-time attacks that inject hidden functionality into a model, which is triggered at inference time in the presence of a specific input pattern. Backdoors are typically realized through \emph{data poisoning}: injecting a subset of malicious samples into the training set so that the model learns the unintended behavior during normal training. Dirty-label backdoors modify both the input (via a trigger) and the label of poisoned samples to a target class chosen by the attacker~\cite{gu2019badnets}. In contrast, clean-label backdoor attacks preserve the original label, making detection more difficult~\cite{turner2018clean}.

\begin{definition}[trigger] \label{def:trigger}
    For a given NN $f: \mathcal{X} \rightarrow \mathcal{Y}$, 
    a distance $\mathrm{d}:\mathcal{X} \times \mathcal{X} \rightarrow \mathbb{R}^+$ and a target label $y_t \in \mathcal{Y}$.
    A \textbf{trigger} is a function $\pi: \mathcal{X} \rightarrow \mathcal{X}$ with the following two properties:
    \[
\begin{aligned}
\text{(i)}\;& \mathrm{d}(\pi(\mathbf{x}), \mathbf{x}) \le \varepsilon_1,\\
\text{(ii)}\;& \mathrm{P}\bigl(f(\pi(\mathbf{x})) = y_t\bigr) \ge 1-\varepsilon_2,
\end{aligned}
\]
    where $\varepsilon_1,\varepsilon_2$ are two small positive constants.
    The quantity $1-\varepsilon_2$ can be estimated \emph{a posteriori} with the \textbf{attack success rate} (ASR) and corresponds to the probability that a triggered input is classified as the target label.
\end{definition}

Blend is an illustrative example of trigger~\cite{chen2017targeted}.
In Blend, the attacker chooses a pattern $\bm \tau \in \mathcal{X}$ to blend with the original input $\bm x \in \mathcal{X}$.
The trigger function $\pi: \mathcal{X} \to \mathcal{X}$ is defined by
$
  \pi(\bm x) = (1 - \alpha) \bm x + \alpha \bm \tau,
  \label{eq:trigger_addition}
$
for some small positive constant $\alpha$.
As $\bm x - \pi(\bm x) = \alpha (\bm x  - \bm \tau)$, each pixel of $\pi(\bm x)$ is at most $\alpha$ far from the original input.
As a consequence 
$
    \mathbf{d}_\infty(\pi(\bm x),\bm x )  \leq \alpha.
$
This agrees with the first condition of Definition~\ref{def:trigger}.
For other trigger types, such as spatial transformations (WaNet)~\cite{nguyen2021wanet}, the transformation $\pi$ follows the specific attack's design, and it is not always possible to describe the distance under which the trigger induces a small perturbation.
Triggers can exploit vulnerabilities at different levels of the model: the input space~\cite{gu2019badnets}, the feature space~\cite{nguyen2021wanet}, or the parameter space~\cite{xugrond}. While backdoor attacks have been studied extensively across multiple domains, including images~\cite{gu2019badnets}, audio~\cite{koffas2022can}, and text~\cite{wang2024badagent}, our work focuses on image classification.
We make a standard assumption about the trigger and the backdoored model:

\begin{assumption}[Robustness]
\label{ass:robustness}  
Let $\bm x \in \mathcal{D}$ such that $f(\pi(\bm x))= y_t$. The trigger is robust in $\bm x$ if there exists $\varepsilon > 0$ such that 
\begin{equation}
  \mathrm{P}\big(f\big(\pi(\bm x) + \bm \delta) = y_t, \|\bm \delta\| \leq \varepsilon \big) \geq 1 - \eta,
\end{equation}
for some small $\eta > 0.$
Similarly, the model is robust in $\bm x \in \mathcal{D}$ if there exists $\varepsilon > 0$ such that 
\begin{equation}
  \mathrm{P}\big(f(\bm x + \bm \delta) = y_\mathrm{clean}, \| \bm \delta\| \leq \varepsilon \big) \geq 1 - \eta,
\end{equation}
for some small $\eta > 0.$
\end{assumption}

Assumption~\ref{ass:robustness} is satisfied by any backdoor or model intended to survive realistic scenarios, where triggers may be subject to compression, resizing, e.g., .jpeg compression~\cite{xue2023compression, ferrari2023compress}, and where using the model on clean data behaves similarly to a clean model.
In Assumption~\ref{ass:robustness}, we deliberately choose not to commit to any distance, as different realistic scenarios will introduce perturbations that are small under a specific metric but might be large in another one.

The assumption of robustness seems incompatible with the existence of a trigger like in Definition~\ref{def:trigger}, which limits the distance between $\bm x$ and $\pi(\bm x)$ to $\varepsilon_1 .$ With such a trigger we would have that $\bm \delta = \pi(\bm x) - \bm x$ is a small perturbation of norm bounded by $\varepsilon_1$, then $f(\pi(\bm x) + \bm \delta) = f(\bm x) = y_{\text{clean}}$ which, for $\varepsilon_1 \leq \varepsilon$, contradicts the robustness of the trigger, while $f(\bm x - \bm \delta) = f(\pi(\bm x))= y_t$ contradicts the robustness of the model. The probabilistic nature of Assumption~\ref{ass:robustness} allows us to circumvent this apparent contradiction. We only require that most of the perturbations will not drive the classification of $\pi(\bm x)$ back to the clean label or any other label than the target label $y_t$. This would be impossible to achieve in a low-dimensional space, but it is not exceptional in a higher-dimensional space.

\section{Threat Model}

We adopt the standard backdoor attack threat model from the literature~\cite{li2022backdoor, abad2025sok}, in which an attacker gains control over either the training data, the training process, or both. We consider three realistic deployment scenarios:

\begin{enumerate}
    \item An attacker poisons a dataset and publishes it on popular repositories such as GitHub or HuggingFace. A victim downloads and trains on this dataset, unknowingly implanting the backdoor. The attacker later activates the malicious behavior by presenting the trigger at inference time.
    
    
    \item A victim possesses a clean dataset and specifies the model architecture, but lacks computational resources for training. The victim outsources training to a third party (the attacker), who injects the backdoor during the training process and returns the compromised model. The attacker retains knowledge of the trigger and can exploit the backdoor once deployed.
\end{enumerate}

We assume the attacker has access to the training dataset and can inject a controlled fraction of poisoned samples. The attacker selects a target label $y_t$ and applies a trigger transformation $\pi$ to poisoned samples, relabeling them to $y_t$ (dirty-label poisoning~\cite{gu2019badnets}). The training process itself remains standard; no modifications to the optimization procedure, loss function, or architecture are required.
Under realistic conditions, a defender does not know which trigger (if any) was used during training. Even in the best-case scenario, where the defender successfully identifies or reconstructs the original trigger, through reverse engineering~\cite{wang2019neural}, anomaly detection~\cite{tran2018spectral}, or unlearning~\cite{li2021neural}.

\section{Method}
\label{sec:method}

\subsection{Backdoor Region}
Let $f = g \circ \varphi$ be a trained classifier, where $\varphi: \mathcal{X} \to \mathcal{Z}$ is the feature extractor (e.g., all layers up to a chosen internal layer $\ell$), and $g: \mathcal{Z} \to \mathcal{Y}$ is the final classifier. We write $\varphi_\ell(\bm x) \in \mathbb{R}^d$ for the feature vector at layer $\ell$ for input $\bm x$, and let $y_t$ denote the backdoor target label. In a model with a backdoor, a trigger transformation $\pi_{\text{orig}}: \mathcal{X} \to \mathcal{X}$ (e.g., adding a patch) is constructed such that, for most clean inputs $\bm x \sim \mathcal{D}$,
\begin{equation}
  f\big(\pi_{\text{orig}}(\bm x)\big) = y_t.
\end{equation}
We interpret this as evidence that $\pi_{\text{orig}}$ maps inputs into a \emph{backdoor region} $R_t \subset \mathcal{Z}$ in feature space associated with the target label:
\begin{equation}
  R_t \;\subseteq\; \{ \bm z \in \mathcal{Z} : g(\bm z) = y_t \}.
\end{equation}

\paragraph{Alternative triggers.}

An \emph{alternative trigger} is any transformation $\pi': \mathcal{X} \to \mathcal{X}$ that does not coincide with $\pi_{\text{orig}}$ in input space but still drives representations into the same backdoor region:
\begin{equation}
  \Pr_{\bm x \sim \mathcal{D}}
  \big[g\!\big(\varphi_\ell(\pi'(\bm x))\big) = y_t\big]
  \;\text{ is large.}
\end{equation}
Our goal is not to prescribe a particular parametric form for $\pi'$, but to \emph{systematically search} for such transformations by guiding the feature representation toward the backdoor region $R_t$.

\subsection{Estimating the Backdoor Direction}
\label{sec:backdoor_direction}

We first extract a \emph{backdoor direction} in feature space that captures how the backdoor trigger modifies internal representations. Given a backdoored model $f$ and dataset $\mathcal{D}$, we construct:
\begin{itemize}
  \item a clean set
    $\mathcal{X}_{\text{clean}} = \{\bm x \in \mathcal{D} :
      f(\bm x) = y(\bm x)\}$ of correctly classified validation samples,
  \item a triggered set
    $\mathcal{X}_{\text{trig}}
      = \{\pi_{\text{orig}}(\bm x) : \bm x \in \mathcal{X}_{\text{clean}}\}$
    where the original trigger is applied.
\end{itemize}
At a chosen layer $\ell$, we compute the mean feature vectors
\begin{equation}
  \bm \mu_\ell^{\text{clean}}
  = \mathbb{E}_{\bm x \in \mathcal{X}_{\text{clean}}}
    \big[\varphi_\ell(\bm x)\big],
  \qquad
  \bm \mu_\ell^{\text{trig}}
  = \mathbb{E}_{\bm x \in \mathcal{X}_{\text{trig}}}
    \big[\varphi_\ell(\bm x)\big],
\end{equation}
and define the (normalized) backdoor direction
\begin{equation}
  \bm d_\ell
  = \frac{\bm \mu_\ell^{\text{trig}} - \bm \mu_\ell^{\text{clean}}}
         {\big\|\bm \mu_\ell^{\text{trig}} - \bm \mu_\ell^{\text{clean}}\big\|_2}.
  \label{eq:backdoor_direction}
\end{equation}
Intuitively, $\bm d_\ell$ captures how the original trigger shifts features from the clean region into the backdoor region $R_t$. This direction is model- and trigger-specific and can be recomputed for any variant of the model (e.g., before and after a defense is applied).





\subsection{Feature-Guided Alternative Trigger Generation}
\label{sec:guided_pgd}

Standard targeted adversarial attacks (e.g.,~\cite{madry2017towards, goodfellow2014explaining}) optimize only the classification loss toward the target label $y_t$:
\begin{equation}
  \min_{\bm \delta} \;
  \mathrm{CE}\big(f(\bm x+\bm \delta), y_t\big)
  \quad \text{s.t.} \quad
  \|\bm \delta\|_\infty \leq \varepsilon,
  \label{eq:standard_targeted}
\end{equation}
where $\varepsilon$ bounds the perturbation under the $\ell_\infty$ norm. Other norms can also be used, but we restrict our attention to $\ell_\infty$, which is the most common in the literature.
Such attacks are free to exploit any vulnerability of the model and are not constrained to follow the backdoor mechanism.

To specifically probe the backdoor region $R_t$, we introduce a \emph{feature-guided} objective that jointly: (i) drives the model toward predicting $y_t$, and (ii) encourages the internal representation to align with the backdoor direction $\bm d_\ell$:
\begin{equation}
  J(\bm x)
  = - \mathrm{CE}\big(f(\bm x), y_t\big)
    + \beta \,\big\langle \varphi_\ell(\bm x),\, \bm d_\ell \big\rangle,
  \label{eq:guided_objective}
\end{equation}
where $\beta > 0$ controls the strength of feature guidance, $\mathrm{CE}$ is the cross-entropy loss, and $\langle \cdot, \cdot \rangle$ denotes the inner product in feature space.

Starting from a clean input $x$, we initialize $\bm x^{(0)} = \bm x + \bm \xi$ with a small random perturbation $\bm \xi \sim_\mathcal{U}[-\eta, \eta]^n$ and perform projected gradient ascent to maximize $J$ under an $\ell_\infty$ constraint:
\begin{align}
  \bm x^{(k+1)}
  &=
  \Pi_{B_\infty(\bm x,\varepsilon)}\!\left(
    \bm x^{(k)} + \alpha \,\mathrm{sign}
      \big(\nabla_{\bm x}  J(\bm x^{(k)})\big)
  \right), \\
  &\hspace{8em} k = 0,\dots,K-1,
  \label{eq:guided_pgd_update}
\end{align}
where $\alpha$ is the step size, $K$ is the number of iterations, $B_\infty(\bm x,\varepsilon)$ is the $\ell_\infty$-ball of radius $\varepsilon$ around $\bm x$, and $\Pi_{B_\infty(\bm x,\varepsilon)}$ denotes projection onto this ball followed by clipping to the valid pixel range $[0,1]$. The final alternative trigger is $\pi'(\bm x)\ = \bm x^{(K)}$.

Because the second term in Eq.~\eqref{eq:guided_objective} explicitly maximizes the projection of $\varphi_\ell(\bm x)$ onto $\bm d_\ell$, our feature-guided attack (FGA) constructs perturbations that not only cause misclassification to $y_t$ but also \emph{use the same internal feature direction} as the original trigger. We treat successful $\pi'(\bm x)$ as \emph{alternative trigger}: they are distinct from the original trigger in input space yet are constructed to exploit the backdoor mechanism encoded by $\bm d_\ell$.

\subsection{Interpretation of Alternative Triggers}
\label{sec:existence}

To understand how alternative triggers work, we start by providing an interpretation of the mechanism underlying the original trigger $\pi$. During poisoned training, the model learns a shortcut: inputs with the trigger are mapped to the target label regardless of their original semantic content.

This creates a \emph{backdoor region} $R_t \subset \mathcal{Z}$ in feature space defined as
\begin{equation}
  R_t = \{ \bm z \in \mathcal{Z} : g(\bm z) = y_t \},
\end{equation}
where $g: \mathcal{Z} \to \mathcal{Y}$ is the classifier head. Any input whose feature representation lies in $R_t$ is classified as the target label $y_t$.
Ideally, if we apply the trigger to the whole dataset, we will have $\varphi(\pi(\mathcal{D})) \subseteq R_t.$
Consider $\bm x \in \mathcal{D}$, if we apply the trigger $\pi$ to it and the feature extractor we will get $\varphi(\pi( \bm  x)) = \varphi(\bm  x) + \bm  d_x$ for some vector $\bm d_x.$
The vector $\bm  d_\ell$ defined in Eq.~\eqref{eq:backdoor_direction} is exactly the expected value of $\bm d_x.$
It is then reasonable to expect that a large fraction of $\bm d_x$ have a large inner product with $\bm d_\ell,$ that is $P(\langle \bm  d_x, \bm d_\ell \rangle \geq \lambda \| \bm d_x \|_2 ) = 1 - \eta$ for some relatively large $0 < \lambda \leq 1 $ and a small $0 < \eta < 1$. In practice, we expect the vector $\bm d_x$ to have a large projection over $\bm d_\ell$ whenever the attack succeeds, i.e., with a probability close to ASR.

Before the final classification, the feature vector is passed through the last layer, which is of the form $\text{softmax}(\bm W^{(\ell)} \varphi(\bm x) + \bm b^{(\ell)})$, where each row of $\bm W^{(\ell)}$ will contribute to the score of a different label.
Experimental results (see Figure \ref{fig:feature_projection}) show that $\bm v = \bm W^{(\ell)} \bm d_\ell$ is a vector with a large positive component $v_t$, where $t$ is the target label's index, and a small component in the other entries.
Let $\bm w^{(\ell)}_t$ be the $t$-th row of $\bm W^{(\ell)}$ corresponding to the target label.
Then, we have $\bm w^{(\ell)}_t = \hat{\bm w}^{(\ell)}_t + v_t \bm d_\ell$ where $\hat{\bm w}^{(\ell)}_t$ lies in the space orthogonal to $\bm d_\ell$.
Our assumption is that $\bm d_\ell^\intercal \bm d_x \geq \lambda \| \bm d_x \|_2$ with a probability close to ASR.
When we plug this into the last layer, we obtain
$
\bm W^{(\ell)} (\varphi(\bm x) + \bm d_x) = \bm W^{(\ell)}\varphi(\bm x) + \bm v_x.
$
The target component will be
$
v_{x,t} = \hat{\bm w}^{(\ell)\intercal}_t \bm d_x + v_t \bm d_{\ell}^\intercal  \bm d_x = r + v_t \lambda
$
where $r = \hat{\bm w}^{(\ell)\intercal}_t \bm d_x$ is expected to be small due to the high dimensionality.

In our assumption, $\lambda$ is non-negligible, and $v_t$ is large, which explains how the trigger will bias the classification in favor of $y_t$.
Any perturbation that induces a similar shift in direction $\bm d_\ell$ in the feature space will increase the chances of classifying the input in the target label, and the loss function (Eq.~\eqref{eq:guided_objective}) pushes exactly in this direction.

An interpretation for the existence of alternative triggers is that, given the contractive nature of the neural network between the input space and the feature space, there are multiple ways to obtain a shift in direction $\bm d_\ell$.
One of these ways is represented by the original trigger $\pi(\cdot)$, while other ways correspond to alternative triggers using the same backdoor.

Geometrically, the points in the dataset $\mathcal{D}$ are mapped to a cloud of points $\varphi(\mathcal{D})$ in the feature space. Once we remove the points legitimately belonging to the target class, this cloud is thin \footnote{By thin we mean that the cloud is relatively flat along the direction $\bm d_\ell$.} in direction $\bm d_\ell,$ the trigger maps the same points in a similar cloud $\varphi(\mathcal{D})$ shifted on average by $\bm d_\ell$.
The trigger removal techniques proposed in Section~\ref{sec:defenses} reduce the distance between these two clouds, making this defense effective for the original trigger $\pi$.

The existence of an alternative trigger that activates the same backdoor in the feature space suggests that the model, after the unlearning process, still maps the data points in a cloud that is thin in direction $\bm d_\ell$, while the new $\bm w^{(\ell)}_t$ still has a large component in $\bm d_\ell$.

\section{Experimentation}

We structure our evaluation into two parts: (i) baseline experiments examining how standard adversarial attacks interact with backdoored models, and (ii) evaluation of our Feature-Guided Attack (FGA) to demonstrate that alternative triggers exist and remain effective even after state-of-the-art defenses are applied. Experimental details can be found in Appendix~\ref{app:exp_details}. Additional experimentation on different hyperparameters is in Appendix~\ref{sec:hyperparams} and on the perceptual similarity in Appendix~\ref{sec:perceptual_sim}.

\subsection{Experimental Setup}

We evaluate our approach on CIFAR-10~\cite{krizhevsky2009learning} ($32\times32\times3$), CIFAR-100~\cite{krizhevsky2009learning} ($32\times32\times3$), and TinyImageNet~\cite{le2015tiny} ($64\times64\times3$), using ResNet-18~\cite{he2016deep} and VGG-19~\cite{simonyan2014very} initialized from ImageNet-pretrained \texttt{torchvision} checkpoints. For each (dataset, architecture) pair, we implant backdoors using four common attack methods: BadNets~\cite{gu2019badnets}, Blend~\cite{li2021invisible}, WaNet~\cite{nguyen2021wanet}, and Input-Aware~\cite{nguyen2020input}, poisoning 5\% or 10\% of training images. We use universal (BadNets and Blended) and sample-specific triggers (WaNet and Input-Aware). Poisoned samples are relabeled to a fixed target class. We use 0 as the target class for simplicity. Models are fine-tuned for up to 20 epochs with early stopping using cross-entropy loss and SGD with mixed precision in PyTorch. All experiments are conducted on a single Linux machine with a 12\,GB NVIDIA GPU and 32\,GB RAM. Results are reported in Table~\ref{tab:backdoor_results} and Figure~\ref{fig:backdoor_results} in Appendix.

\subsection{Baseline: Standard Adversarial Attacks on Backdoored Models}

We first ask whether standard adversarial attacks can naturally exploit the shortcut created by backdoor training. Specifically, we evaluate: (i) untargeted PGD, to determine whether misclassifications under adversarial noise preferentially select the backdoor target class, and (ii) targeted PGD toward the backdoor target label, as an upper bound for pixel-space optimization.

\paragraph{Clean model baselines.}
Clean (unbackdoored) models achieve validation accuracies of 79.04\% and 88.14\% on CIFAR-10, 52.40\% and 63.89\% on CIFAR-100, and 53.96\% and 61.56\% on TinyImageNet, for ResNet-18 and VGG-19, respectively. Applying the trigger pattern yields ASR close to random chance (approximately 10\% on CIFAR-10, 1\% on CIFAR-100, and 0.5\% on TinyImageNet), confirming that the models have no inherent backdoor region. 

\begin{figure*}[htb]
  \centering
  \begin{subfigure}[t]{0.23\textwidth}
    \centering
    \includegraphics[width=\linewidth]{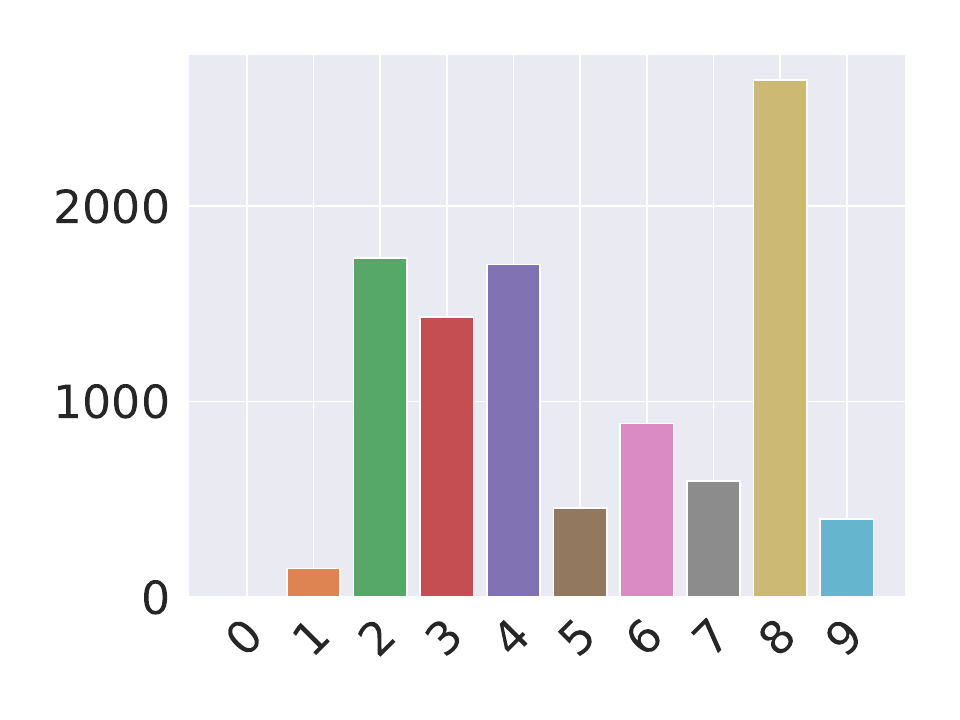}
    \caption{BadNets.}
    \label{fig:distribution_badnets}
  \end{subfigure}
  \begin{subfigure}[t]{0.23\textwidth}
    \centering
    \includegraphics[width=\linewidth]{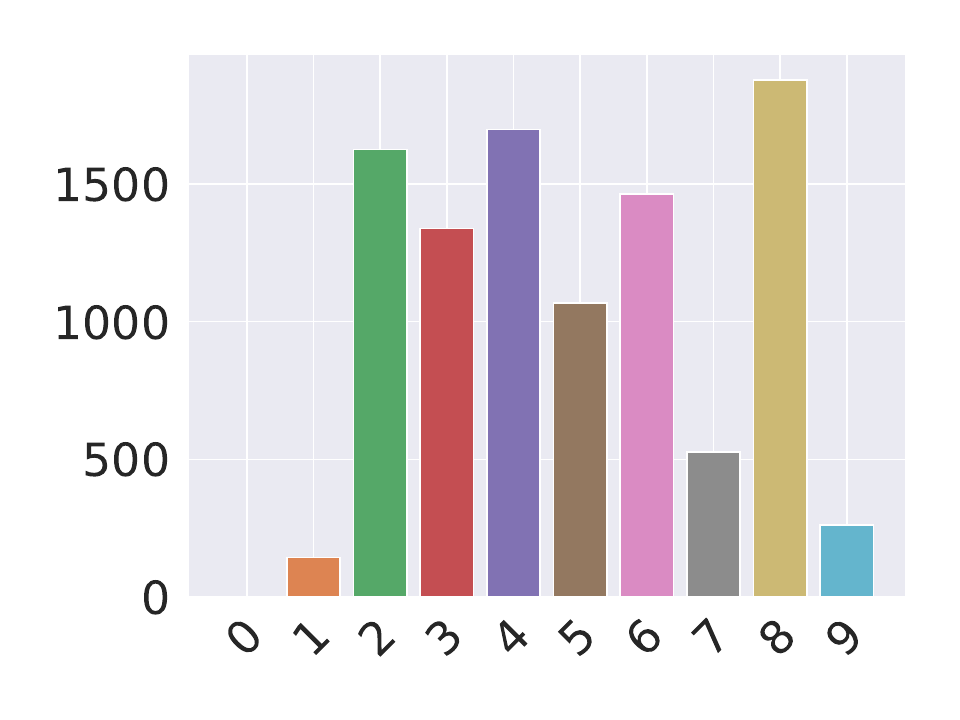}
    \caption{Blend.}
    \label{fig:distribution_blend}
  \end{subfigure}
  \begin{subfigure}[t]{0.23\textwidth}
    \centering
    \includegraphics[width=\linewidth]{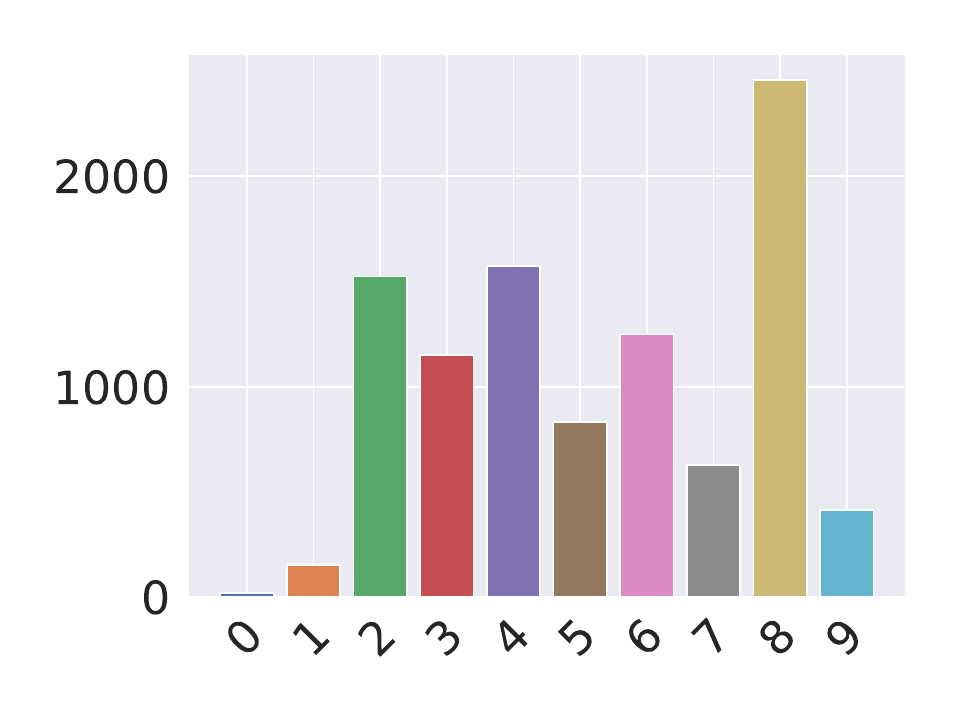}
    \caption{WaNet.}
    \label{fig:distribution_wanet}
  \end{subfigure}
\begin{subfigure}[t]{0.23\textwidth}
    \centering
    \includegraphics[width=\linewidth]{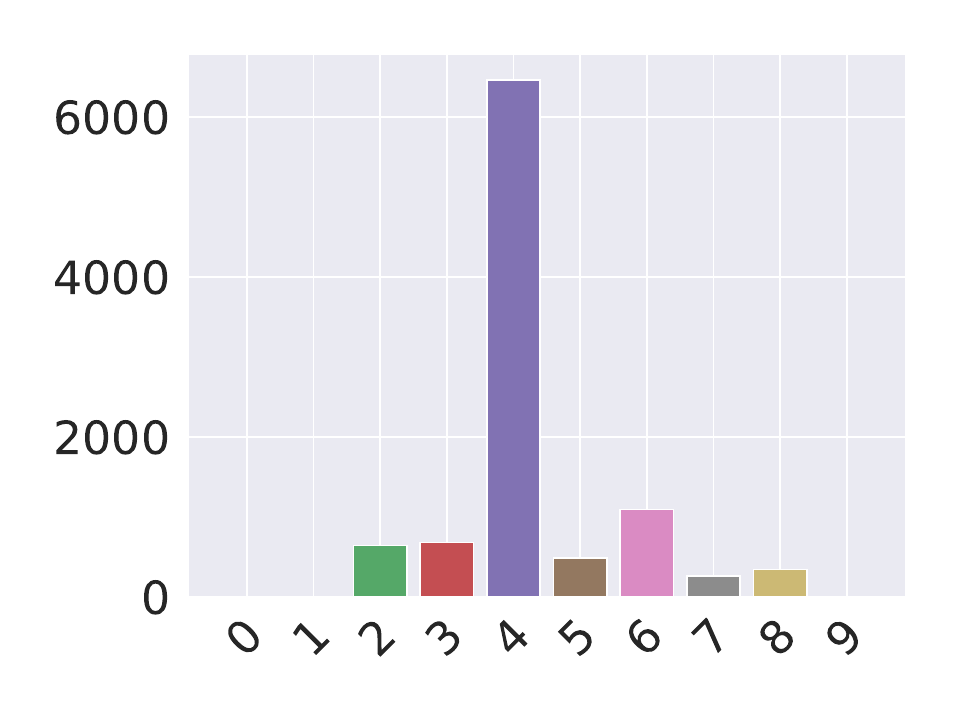}
    \caption{Input-Aware.}
    \label{fig:distribution_inputaware}
  \end{subfigure}  
  \caption{Distribution of predicted labels under untargeted PGD on backdoored ResNet-18 (CIFAR-10) with different trigger types. We observe a spread distribution and no bias towards the target label. The X-axis is the class label, and the Y-axis is the count of samples.}
  \label{fig:distribution_backdoors}
\end{figure*}

\paragraph{Untargeted PGD on backdoored models.}
After injecting the backdoor (target label 0), untargeted PGD achieves high overall misclassification rates but does not preferentially predict the backdoor target class (Figure~\ref{fig:distribution_backdoors}). Notably, the target label is among the \emph{least} selected classes. This is because untargeted PGD minimizes the loss of the correct class with no constraint on which alternative class is selected. It therefore crosses the nearest decision boundary. The backdoor region is persistent but not near for most inputs. This means that the absence of adversarial examples targeting the backdoor class cannot be taken as evidence that the backdoor is inaccessible or non-existent.
    
\paragraph{Targeted PGD and the existence of alternative triggers.}

A natural question arises: can standard targeted adversarial attacks discover alternative triggers without explicit feature-space guidance? Table~\ref{tab:tpgd_results} shows that targeted PGD (T-PGD) toward the backdoor label achieves high success rates in many configurations. This demonstrates that alternative triggers are not rare. Rather, they emerge naturally from the optimization landscape created by backdoor training.

However, high ASR alone does not guarantee exploitation of the backdoor mechanism. Targeted attacks optimize solely for predicting the target class. They may use the backdoor region, or they may find entirely different adversarial shortcuts unrelated to the backdoor. Without explicit constraints on the feature-space path, we cannot determine whether successful perturbations are functionally equivalent to the original trigger or simply arbitrary adversarial examples that happen to reach the target class. For example, an image of a ``cat'' can be misclassified as a ``dog'' by either exploiting the backdoor space generated by the trigger or using ``dog'' latent space. FGA ensures that the backdoor space is used. This observation motivates FGA's design. By explicitly maximizing alignment with the estimated backdoor direction $\bm d_\ell$ (Eq.~\eqref{eq:guided_objective}), we ensure that discovered triggers exploit the same latent vulnerability as the original backdoor.

\begin{table}[htb]
\small
\centering
\caption{T-PGD ASR (\%) toward the backdoor target on backdoored ResNet-18 models.
  T-PGD optimizes solely for the target class without feature-space guidance. It achieves high ASR on CIFAR-10/100 at larger $\varepsilon$ but struggles on TinyImageNet at small budgets, suggesting feature guidance gains value as input dimensionality grows.}
\label{tab:tpgd_results}
\begin{tabular}{lc cccc}
\toprule
Dataset & $\varepsilon$ & BadNets & Blend & WaNet & Input-Aware \\
\midrule
\multirow{3}{*}{CIFAR-10}
  & $8/255$  & 98.18 & 96.99 & 99.07 & 22.34 \\
  & $16/255$ & 99.73 & 99.09 & 99.96 & 64.13 \\
  & $32/255$ & 99.89 & 99.67 & 100.0 & 93.63 \\
\midrule
\multirow{3}{*}{CIFAR-100}
  & $8/255$  & 75.19 & 72.90 & 83.05 &  7.14 \\
  & $16/255$ & 94.27 & 89.06 & 96.47 & 43.53 \\
  & $32/255$ & 98.37 & 95.48 & 99.17 & 82.09 \\
\midrule
\multirow{3}{*}{TinyImageNet}
  & $8/255$  & 34.49 & 91.50 & 50.56 & 10.70 \\
  & $16/255$ & 73.22 & 95.74 & 83.37 & 32.52 \\
  & $32/255$ & 93.20 & 98.08 & 96.36 & 68.80 \\
\bottomrule
\end{tabular}
\end{table}


\subsection{Validating the Backdoor Direction}

To verify that the estimated direction $\bm d_\ell$ accurately captures the backdoor mechanism, we perform a feature-space interpolation experiment. For a set of correctly-classified clean samples, we compute their feature representations $\varphi(\bm x_{\text{clean}})$ and the corresponding triggered representations $\varphi(\bm x_{\text{trig}})$. We then evaluate the model's predictions along the path connecting these two points in feature space:
\begin{equation}
\varphi_\alpha = \varphi(\bm x_{\text{clean}}) + \alpha \bm  d_\ell,
\end{equation}
where $s = \|\varphi(\bm  x_{\text{trig}}) - \varphi(\bm x_{\text{clean}})\|$ is the displacement magnitude for that sample, and $\alpha \in [0, 1.5]$ controls the interpolation. At $\alpha=0$, we have the clean feature, and at $\alpha=1$, we reach the triggered feature.

\begin{figure}[htb]
    \centering
    \includegraphics[width=0.8\textwidth]{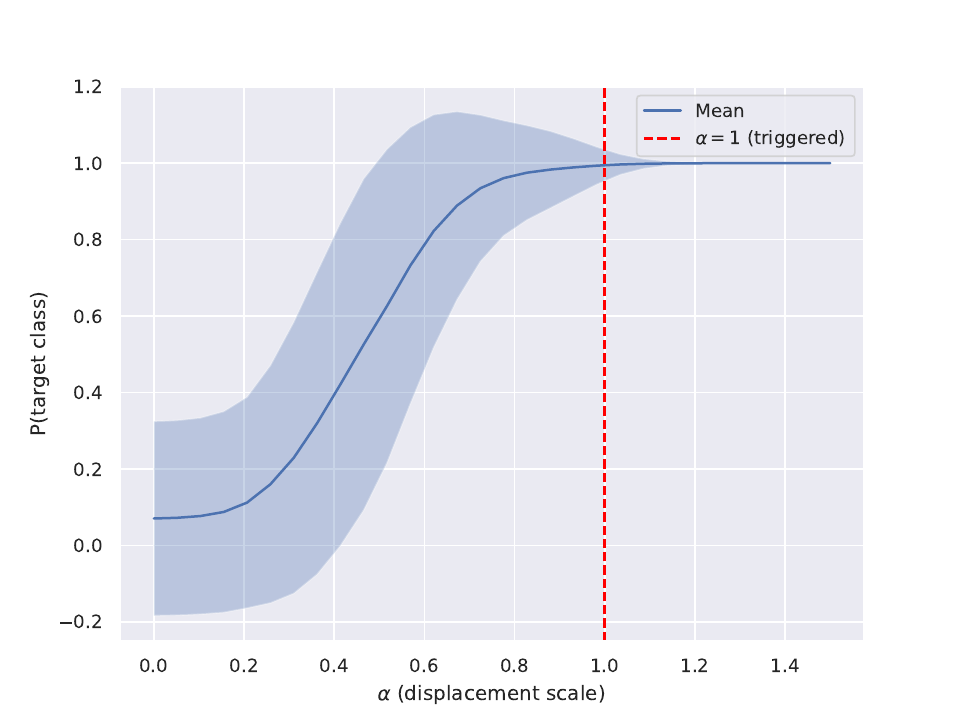}
    \caption{Target class probability as a function of displacement $\alpha$ along the estimated backdoor direction $\bm d_\ell$ on CIFAR-10 (ResNet-18, BadNets). The solid line shows the mean over 100 samples. At $\alpha=1$ (red dashed line), features match the triggered samples and achieve near-perfect target classification.}
    \label{fig:feature_projection}
\end{figure}

Figure~\ref{fig:feature_projection} shows the target class probability as a function of $\alpha$, averaged over 100 randomly selected CIFAR-10 samples (ResNet-18, BadNets, 10\% poison rate). At $\alpha=0$, the mean target probability is approximately 7\%, close to the random baseline for a 10-class problem. As $\alpha$ increases, the probability rises smoothly in a sigmoid-like transition, reaching approximately 90\% by $\alpha=0.6$ and saturating near 100\% by $\alpha=1.0$. Beyond $\alpha=1$, the probability remains stable, indicating that the backdoor is fully activated once features reach the triggered region. The smooth transition demonstrates two key properties of our estimated backdoor direction: (i) $\bm d_\ell$ is \emph{aligned} with the actual feature-space displacement induced by the backdoor trigger, and (ii) the backdoor region is \emph{concentrated}, because small steps along $\bm d_\ell$ are sufficient to flip predictions. The high variance in the transition region ($\alpha \approx 0.3$-$0.7$) reflects sample-to-sample variability in how strongly the backdoor affects different inputs, but all samples ultimately converge to high target probability by $\alpha=1$. This experiment provides empirical evidence that $\bm d_\ell$ is not merely a heuristic but a representation of the latent mechanism by which backdoors manipulate model predictions.

\subsection{Discovering Alternative Triggers}

After observing that the backdoor region can be observed by feature-guided optimization, here, we analyze the performance of FGA under different settings. Table~\ref{tab:backdoor_results} reports the FGA success rate toward the backdoor target label across all settings. We use 200 optimization steps, step size $\alpha = 2/255$, feature guidance weight $\beta = 1$, and a maximum $\ell_\infty$ perturbation budget of $\varepsilon = 32/255$. Additional hyperparameter and poisoning rate experiments are reported in Sections~\ref{sec:hyperparams} and~\ref{sec:defenses}.
FGA achieves consistently high ASR across all datasets, architectures, and trigger types. Crucially, this holds even in settings where untargeted PGD fails to preferentially select the target class. Specifically for WaNet on CIFAR-10 (ResNet-18), untargeted PGD is only 1.03\% on the target class, while FGA reaches 100\%. This gap highlights the core advantage of our approach, while untargeted PGD searches indiscriminately over all possible misclassifications, FGA focuses gradient updates on the specific latent subspace activated by the backdoor, enabling the discovery of alternative triggers that are functionally equivalent to the original trigger yet perceptually distinct.

\begin{table*}[!htb]
\small
\centering
\setlength{\tabcolsep}{4pt}   
\caption{
  Validation performance of backdoored models (10\% poison rate) with clean baselines.
  \textbf{Acc}: clean validation accuracy. \textbf{ASR}: backdoor attack success rate.
  \textbf{FGA}: our feature-guided attack success rate.
}
\label{tab:backdoor_results}
\resizebox{\textwidth}{!}{%
\begin{tabular}{llc cccc cccc cccc cccc}
\toprule
 & & & \multicolumn{2}{c}{BadNets} & \multicolumn{2}{c}{Blend}
   & \multicolumn{2}{c}{WaNet}   & \multicolumn{2}{c}{Input-Aware} \\
\cmidrule(lr){4-5}\cmidrule(lr){6-7}\cmidrule(lr){8-9}\cmidrule(lr){10-11}
Dataset & Model & Acc & ASR & \textbf{FGA} & ASR & \textbf{FGA}
                         & ASR & \textbf{FGA} & ASR & \textbf{FGA} \\
\midrule
CIFAR-10  & ResNet-18 & 78.90 & 97.30 & 99.78 & 83.98 & 99.53 & 60.29 & 100.0  & 88.32 & 92.36 \\
CIFAR-10  & VGG-19    & 87.80 & 98.23 & 99.34 & 87.08 & 99.76 & 63.80 &  99.99 & 91.11 & 97.98 \\
CIFAR-100 & ResNet-18 & 51.64 & 94.79 & 96.32 & 81.17 & 94.19 & 53.91 &  98.66 & 85.56 & 79.96 \\
CIFAR-100 & VGG-19    & 62.21 & 95.41 & 96.22 & 90.74 & 95.27 & 55.14 &  99.68 & 91.58 & 95.72 \\
TinyImageNet   & ResNet-18 & 53.34 & 99.64 & 89.48 & 88.00 & 97.52 & 86.82 &  96.30 & 96.42 & 63.44 \\
TinyImageNet   & VGG-19    & 60.84 & 99.86 & 89.96 & 80.98 & 99.06 & 75.50 &  89.38 & 94.08 & 70.82 \\
\bottomrule
\end{tabular}%
}
\end{table*}

\subsection{Defenses}
\label{sec:defenses}

Backdoor defenses can be deployed at different stages of the model pipeline, including pre-training, in-training, and post-training phases~\cite{abad2025sok}. Pre-training defenses aim to sanitize or curate the training data before learning starts, in-training defenses monitor and regularize the optimization process, and post-training defenses analyze or repair an already trained model~\cite{abad2025sok}. Because evaluating alternative triggers assumes access to a backdoored model, this work focuses exclusively on post-training defenses.

Input-space filtering defenses, which detect or suppress specific trigger patterns at inference time, are ineffective against alternative triggers by design, since such triggers can differ substantially in pixel space from the original pattern. For example, many defenses are developed and evaluated against a fixed BadNets-style patch and measure success only by suppressing the response to that exact trigger. Consequently, they effectively break the association between the original BadNets trigger and the backdoor behavior, but provide no guarantee against semantically or functionally equivalent variations of the trigger, i.e., alternative triggers. We therefore restrict our evaluation to defenses that directly repair the model rather than those that operate purely in the input space, e.g.,~\cite{chen2018detecting}.

\paragraph{BAN.}
We evaluate BAN~\cite{xu2024ban} (NeurIPS 2024), a recent post-training defense that repairs the latent space by fine-tuning under adversarial neuron perturbations. The intuition is that backdoor neurons selectively mislead predictions toward the target class under noise, while benign neurons degrade uniformly. BAN suppresses this asymmetry during fine-tuning. As shown in Table~\ref{tab:ban_results} in Appendix~\ref{sec:app-ban}, BAN successfully reduces the original trigger's ASR to near-random levels (8-21\% on CIFAR-10). However, FGA still achieves 74-87\% ASR on the repaired models. 

\paragraph{NAD.}
Neural Attention Distillation (NAD)~\cite{li2021neural} fine-tunes the backdoored model using a clean teacher network, aligning intermediate-layer attention maps to suppress trigger-specific activations. Table~\ref{tab:nad_results} in Appendix~\ref{sec:app-nad} shows that while NAD reduces the original ASR to near-random levels (7-10\% across attack types), FGA continues to succeed at 63-85\% ASR on the repaired models.

\paragraph{Trigger-Aware Unlearning.}

We evaluate a realistic post-training defense where the defender has recovered the original trigger and attempts to neutralize it via fine-tuning. Concretely, we perform \emph{trigger-aware unlearning} by fine-tuning the backdoored model on trigger-stamped images relabeled with their correct classes, assuming the defender has access to (or has reverse-engineered) the trigger pattern, which is plausible given existing methods~\cite{wang2019neural,liu2019abs}. This procedure drives the ASR of the original trigger close to random guessing (Figures~\ref{fig:unlearning_005} and~\ref{fig:unlearning_01} in Appendix~\ref{sec:trigger-aware-unlearning}), while largely preserving clean accuracy, indicating that the trigger–label association is effectively neutralized. However, for Input-Aware triggers, unlearning fails to fully remove the backdoor, likely due to their sample-specific nature. Moreover, previously generated FGA-based alternative triggers often remain highly effective after unlearning, and re-optimizing FGA triggers on the unlearned model again yields strong attacks, see Figure~\ref{fig:difference_asr} in the appendix.

\subsection{Do Alternative Triggers Remove the Backdoor?}

A natural question arises: if unlearning with the original trigger fails to remove the backdoor, what happens if we unlearn using FGA-generated alternative triggers instead? To investigate this, we perform the following experiment: (i) generate alternative triggers using FGA on a backdoored model, (ii) apply trigger-aware unlearning by fine-tuning on these alternative triggers relabeled with their ground-truth classes, and (iii) re-run FGA to search for new alternative triggers on the unlearned model.

Tables~\ref{tab:unlearning_resnet18} and~\ref{tab:unlearning_vgg19} in Appendix~\ref{sec:alt-trigger-as-def} report results for ResNet-18 and VGG-19, respectively. The \emph{ASR after Unlearning} column measures how well the specific FGA trigger used for unlearning is suppressed, while the FGA columns measure the ASR of newly generated alternative triggers on the unlearned model.

Unlearning proves largely ineffective at removing the original trigger (e.g., BadNets) it was trained on. This is because FGA probes a broader backdoor subspace, so suppressing one alternative trigger leaves the subspace intact for FGA to find another. On CIFAR-10, ASR after unlearning remains above 95\% for BadNets, WaNet, and Blend on both architectures, indicating the unlearning procedure fails to neutralize the alternative trigger. Input-Aware is a partial exception, where ASR after unlearning drops to near-random levels, though it remains high on CIFAR-100. This inconsistency across datasets and attacks suggests the unreliability of trigger-aware unlearning as a defense.

\section{Conclusion \& Future Work}
Backdoor training introduces some weakness in the model, allowing the small perturbation represented by the trigger to cause misclassifications towards a target label. This weakness is not restricted to the immediate neighbors of the trigger, but can also be exploited by FGA.

This has two concrete consequences for defenders. First, backdoor detection is easier than previously assumed: a defender need not recover the exact original trigger, only any perturbation that activates the backdoor region. Second, neutralizing a single trigger is insufficient. Standard defenses unlearn the known trigger but leave the backdoor region intact. Alternative triggers remain effective even after the original trigger is reduced to random-chance ASR. Effective defense must target the backdoor region itself, at the latent-space level rather than only at the pixel level.

We formalize these observations theoretically and support them empirically across multiple datasets, architectures, attacks, and defenses. We introduce FGA, a feature-guided attack that constrains perturbations to activate the backdoor region via the same latent direction as the original trigger.

Our alternative triggers are currently sample-specific, optimized per input and model. A natural extension is to study their \emph{transferability}: whether triggers crafted for one model activate the backdoor region of another, as adversarial examples are known to do~\cite{tramer2017space}. A further direction is the construction of \emph{universal} alternative triggers, a single perturbation effective across diverse inputs and models, which would constitute the strongest evidence that the backdoor region is a stable model-level vulnerability. We leave these directions for future work.

\bibliographystyle{plain}
\bibliography{references}

\newpage
\appendix
\section{Notation}
\label{sec:notation}

Table~\ref{tab:notation} provides the notation used throughout the paper.

\begin{table}[htb]
\small
    \caption{List of Symbols}
    \label{tab:notation}
    \centering
    \renewcommand{\arraystretch}{1.1}
    \begin{tabular}{@{} l p{0.7\columnwidth} @{}}
        \hline
        \textbf{Symbol} & \textbf{Description} \\
        \hline\hline
        $\mathbb{R}$ & Set of real numbers \\
        $n$ & Dimensionality of input space / a natural number \\
        $d$ & Dimensionality of feature space \\
        $m$ & Number of classes \\
        $\bm x \in \mathbb{R}^n$ & vectors lower case in bold regardless of the dimension \\
        $\mathbf{0}$ & Zero vector or matrix \\
        $\|\bm x\|_p = \mathrm{d}_p(\bm x, \bm 0)$ & $\ell_p$ distance from $\bm 0$ vector\\
        $\bm W$ & Matrices upper case bold \\
        $\mathcal{X} \subseteq \mathbb{R}^n$ & Set of all possible inputs \\
        $\mathcal{Z} \subseteq \mathbb{R}^d$ & Set of all possible features \\
        $\mathcal{Y}$ & Set of all possible labels \\
        $\mathcal{D} \subseteq \mathcal{X} \times \mathcal{Y}$ & Training dataset \\
        $y_t \in \mathcal{Y}$ & Target label\\
        $\varepsilon, \eta$ & Small positive quantities \\
        $f(\cdot)$ & General function mapping inputs to outputs \\
        $\varphi(\cdot)$ & Feature extractor \\        
        $\pi, \pi_{\text{orig}}, \pi'$ & Trigger function\\
    \hline
    \end{tabular}
\end{table}

\section{Experimental Details}
\label{app:exp_details}

\paragraph{Datasets.}
We use three standard image classification benchmarks. CIFAR-10~\cite{krizhevsky2009learning} is a 10-class dataset with 50\,000 training images and 10\,000 test images of size $32\times32\times3$. CIFAR-100~\cite{krizhevsky2009learning} contains 100 classes with the same image count and dimensions. TinyImageNet~\cite{le2015tiny} consists of 200 classes with 100\,000 training images (500 per class) and 10\,000 test images of size $64\times64\times3$. 

\paragraph{Model architectures.}
We evaluate two widely-used convolutional architectures: ResNet-18~\cite{he2016deep} and VGG-19~\cite{simonyan2014very}. Both models are initialized from ImageNet-pretrained checkpoints provided by PyTorch's \texttt{torchvision} library. For CIFAR experiments, we replace the final fully-connected layer to match the number of classes (10 or 100). For TinyImageNet, we use 200 output classes. The feature extractor $\varphi_\ell$ is defined as the network up to and including the global average pooling layer (layer \texttt{avgpool} in PyTorch), yielding 512-dimensional features for ResNet-18 and VGG-19.

\paragraph{Backdoor attacks.}
We implement four standard backdoor attacks, based on common hyperparameter selection in the literature.

\begin{itemize}
    \item \textbf{BadNets}~\cite{gu2019badnets}: A white square patch of size $3\times3$ pixels placed at the bottom-right corner of the image.
    \item \textbf{Blend}~\cite{chen2017targeted}: A Hello Kitty image blended with the input using a blending coefficient $\alpha = 0.2$.
    \item \textbf{WaNet}~\cite{nguyen2021wanet}: A sample-specific warping-based trigger generated using a control grid of size $4\times4$ and noise magnitude $s = 0.5$.
    \item \textbf{Input-Aware}~\cite{nguyen2020input}: An input-dependent trigger generated by a U-Net encoder-decoder with 64 base filters, trained jointly with the backdoored classifier.
\end{itemize}

For each attack, we poison 5\% and 10\% of the training set by randomly selecting images, applying the trigger transformation $\pi_{\text{orig}}$, and relabeling them to target class 0. The clean validation and test sets remain unmodified.

\paragraph{Training procedure.}
We use standard training procedures~\cite{he2016deep}. All models are fine-tuned using stochastic gradient descent (SGD) with momentum 0.9, weight decay $5\times10^{-4}$, and an initial learning rate of 0.01, decayed by a factor of 0.1 at epochs 10 and 15. We use a batch size of 128 and train for up to 20 epochs with early stopping based on validation accuracy. Training is performed with mixed precision using PyTorch's automatic mixed precision (AMP) to reduce memory consumption. Cross-entropy loss is used for all experiments.

\begin{figure}[htb]
    \centering
    \includegraphics[width=\linewidth]{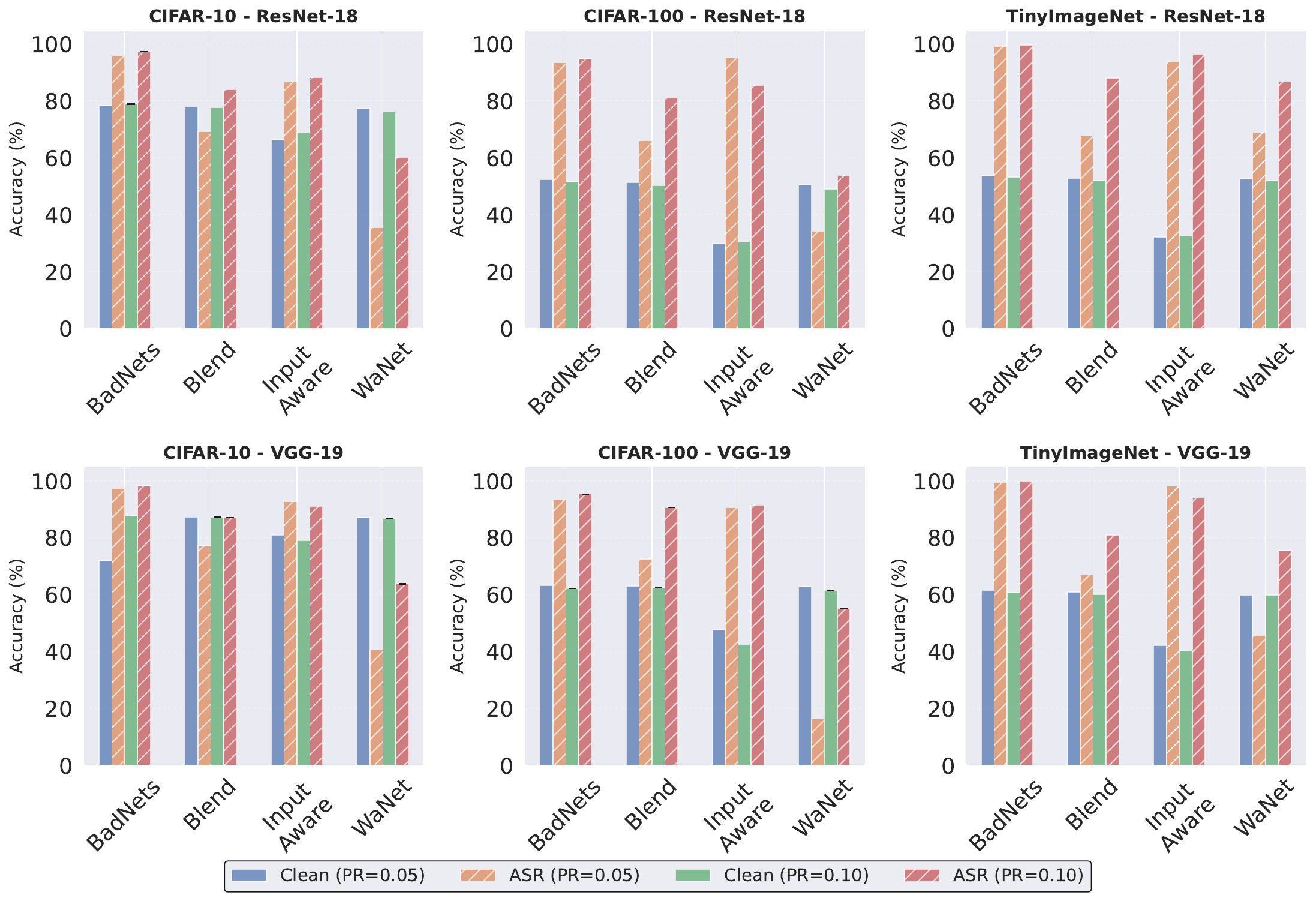}
    \caption{Backdoor performance (clean accuracy and attack success rate) for BadNets, Blend, WaNet, and Input-Aware on CIFAR-10, CIFAR-100, and TinyImageNet using ResNet-18 and VGG-19 under 5\% and 10\% poisoning rates.}
    \label{fig:backdoor_results}
\end{figure}

\paragraph{Backdoor Results.}
Figure~\ref{fig:backdoor_results} shows the backdoor performance for different backdoor attacks under different settings.

\paragraph{Adversarial attack baselines.}
For untargeted PGD~\cite{madry2017towards}, we use 20 iterations with step size $\alpha = 2/255$ and perturbation budget $\varepsilon = 8/255$ under the $\ell_\infty$ norm. For targeted PGD toward the backdoor label, we use the same hyperparameters with the cross-entropy loss modified to encourage prediction of the target class.

\paragraph{FGA hyperparameters.}
Unless otherwise specified, our feature-guided attack (FGA) uses 200 optimization steps with step size $\alpha = 2/255$, feature guidance weight $\beta = 1$, and perturbation budget $\varepsilon = 8/255$. These are common parameters in PGD~\cite{madry2017towards}.

\paragraph{Defense implementations.}
For BAN~\cite{xu2024ban}, we use the official implementation with default hyperparameters: 10 fine-tuning epochs, adversarial neuron noise magnitude $\delta = 0.1$, and a clean fine-tuning set of 5\% of the training data. For NAD~\cite{li2021neural}, we train a clean teacher model on the unpoisoned dataset and fine-tune the backdoored student for 10 epochs with attention alignment weight $\lambda = 0.5$. For trigger-aware unlearning, we fine-tune the backdoored model for 5 epochs on a dataset containing 10\% triggered samples (with correct labels), using the same SGD hyperparameters as initial training.

\paragraph{Evaluation metrics.}
We report clean accuracy (CA) on the unmodified test set and attack success rate (ASR) on the poisoned test set, where ASR is the percentage of triggered samples that are classified as the target label. For defense evaluations, we measure both the ASR of the original trigger and the ASR of FGA-generated alternative triggers.

\section{Results on Defenses}
\label{sec:app_defenses}

\subsection{BAN}
\label{sec:app-ban}

Table~\ref{tab:ban_results} shows the results of BAN against different backdoor attacks and the performance of FGA after the defense has been applied.

\begin{table}[!htb]
\small
\centering
\caption{
Effect of BAN on different backdoored models.
\textbf{ASR after BAN} is the attack success rate of the original trigger
after applying BAN. \textbf{FGA} is the attack success rate of the
alternative triggers on the BAN-cleaned model.
}
\label{tab:ban_results}
\begin{tabular}{ccccc}
\toprule
Dataset & Model    & Attack  & ASR after BAN & \textbf{FGA (Ours)} \\
\midrule
CIFAR-10 & ResNet-18 & BadNets & 11.73 & 80.88 \\
CIFAR-10 & ResNet-18 & WaNet   &  8.58 & 81.96 \\
CIFAR-10 & ResNet-18 & Blend   &  9.44 & 87.48 \\
CIFAR-10 & ResNet-18 & Input-Aware  & 21.13 & 74.22 \\
\bottomrule
\end{tabular}
\end{table}

\subsection{NAD}
\label{sec:app-nad}

Table~\ref{tab:nad_results} shows the results of NAD against different backdoor attacks and the performance of FGA after the defense has been applied.

\begin{table}[!htb]
\small
\centering
\caption{
Effect of NAD on different backdoored models.
\textbf{ASR after NAD} is the attack success rate of the original trigger
after applying NAD. \textbf{FGA} is the attack success rate of the alternative triggers attack on the NAD-cleaned model.
}
\label{tab:nad_results}
\begin{tabular}{ccccc}
\toprule
Dataset & Model    & Attack  & ASR after NAD & \textbf{FGA (Ours)} \\
\midrule
CIFAR-10 & ResNet-18 & BadNets & 8.92 & 63.38 \\
CIFAR-10 & ResNet-18 & WaNet   & 10.37 & 68.06 \\
CIFAR-10 & ResNet-18 & Blend   & 7.86 &  79.11\\
CIFAR-10 & ResNet-18 & Input-Aware & 10.43 & 85.02 \\
\bottomrule
\end{tabular}
\end{table}

\subsection{Trigger Aware Unlearning}
\label{sec:trigger-aware-unlearning}

\begin{figure}[!htb]
    \centering
    \includegraphics[width=\linewidth]{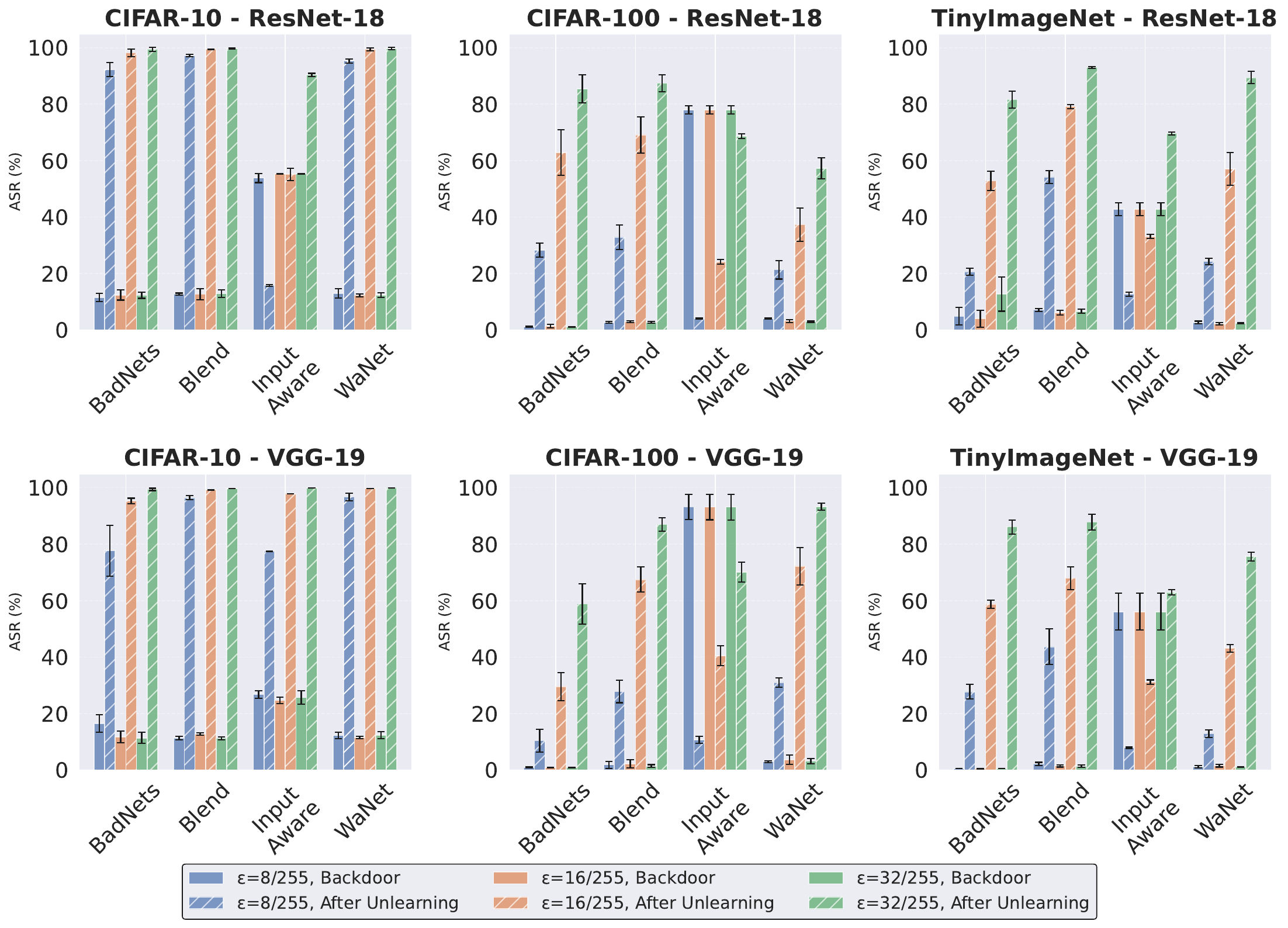}
    \caption{ASR before and after trigger-aware unlearning (5\% poisoning rate). 
    Solid bars show backdoored models; striped bars show post-unlearning results. 
    While unlearning reduces the original trigger ASR to near-random levels, FGA-generated alternative triggers remain highly effective, particularly at larger $\varepsilon$.}
    \label{fig:unlearning_005}
\end{figure}

\begin{figure}[!htb]
    \centering
    \includegraphics[width=\linewidth]{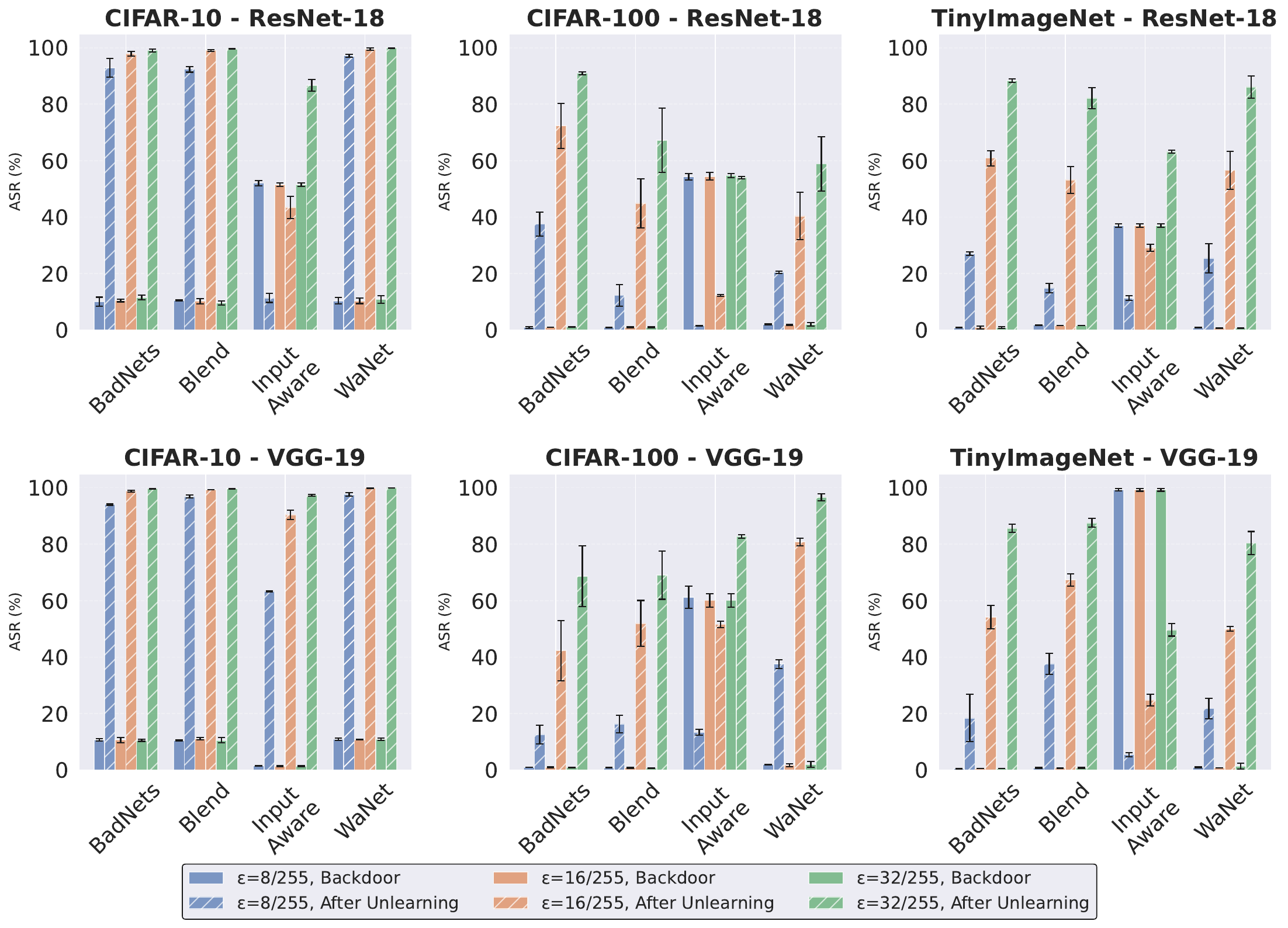}
    \caption{ASR before and after trigger-aware unlearning (10\% poisoning rate). 
    Despite the successful removal of the original trigger, alternative triggers continue to exploit the backdoor feature space, demonstrating that trigger-centric defenses do not fully erase the underlying backdoor mechanism.}
    \label{fig:unlearning_01}
\end{figure}

\begin{figure*}[!htb]
  \centering
  \begin{subfigure}{0.48\textwidth}
    \centering
    \includegraphics[width=\linewidth]{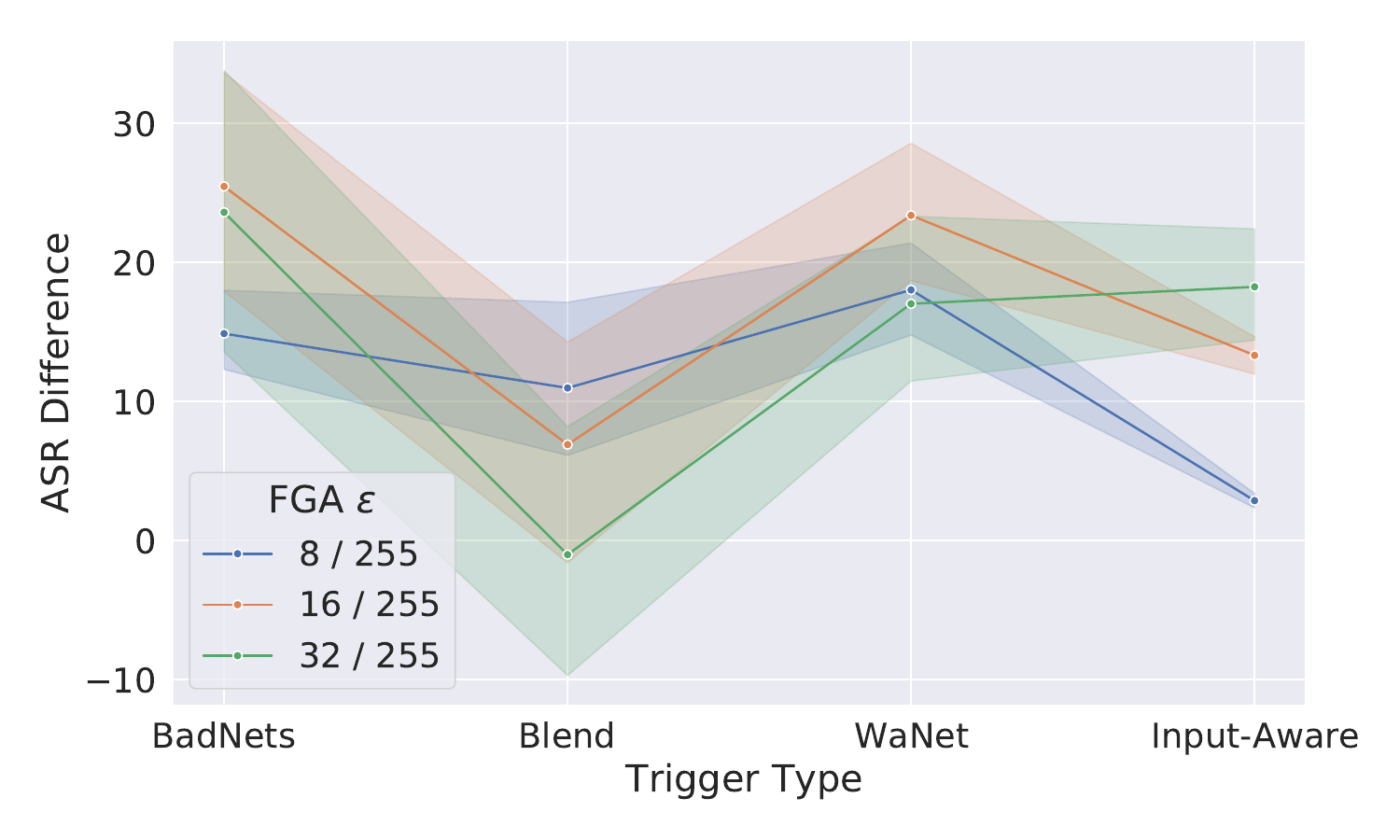}
    \caption{ResNet-18}
  \end{subfigure}
  \hfill
  \begin{subfigure}{0.49\textwidth}
    \centering
    \includegraphics[width=\linewidth]{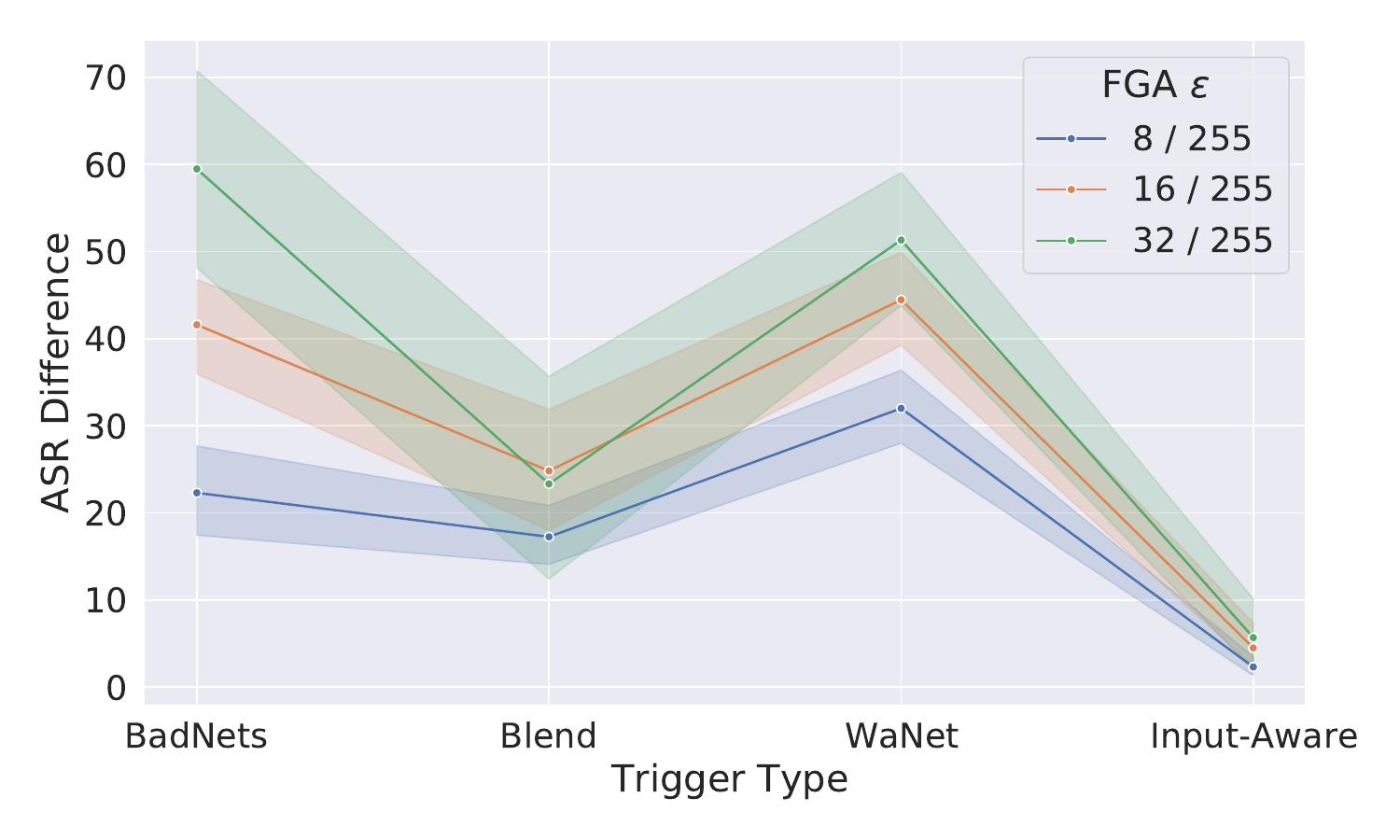}
    \caption{VGG-19}
  \end{subfigure}
    \caption{Change in ASR of FGA-generated alternative triggers before and after trigger-aware unlearning, for different trigger types, datasets, and budgets. Positive values indicate that re-optimizing FGA on the unlearned model yields stronger alternative triggers.}
    \label{fig:difference_asr}
\end{figure*}

Figures~\ref{fig:unlearning_005} and~\ref{fig:unlearning_01} show the performance of the trigger-aware unlearning procedure for various setups. Figure~\ref{fig:difference_asr} shows the difference in ASR for the FGA-generated alternative triggers before and after trigger-aware unlearning

\subsection{Alternative Trigger as a Defense}
\label{sec:alt-trigger-as-def}

Tables~\ref{tab:unlearning_resnet18} and~\ref{tab:unlearning_vgg19} shows FGA re-attack after trigger-aware unlearning on ResNet-18 and VGG-19 respectively across CIFAR-10, CIFAR-100, and TinyImageNet.

\begin{table}[!htb]
\centering
\small
\caption{FGA re-attack after trigger-aware unlearning on \textbf{ResNet-18} across CIFAR-10, CIFAR-100, and TinyImageNet. \emph{ASR after Unlearning} is the attack success rate of the FGA-generated alternative trigger ($\varepsilon=8/255$) after the model has been unlearned on that exact trigger.}
\begin{tabular}{llccccc}
\toprule
Dataset & Attack & \shortstack{ASR after\\Unlearning} & \shortstack{FGA\\$\varepsilon{=}8/255$} & \shortstack{FGA\\$\varepsilon{=}16/255$} & \shortstack{FGA\\$\varepsilon{=}32/255$} \\
\midrule
\multirow{4}{*}{CIFAR-10}     & BadNets     & 97.70 & 97.24 & 99.46 & 99.75  \\
                               & WaNet       & 99.28 & 99.04 & 99.97 & 100.00 \\
                               & Blend       & 95.73 & 95.13 & 98.47 & 99.49  \\
                               & Input-Aware & 17.76 & 10.39 & 23.53 & 73.75  \\
\midrule
\multirow{4}{*}{CIFAR-100}    & BadNets     & 73.35 & 69.32 & 91.71 & 97.41  \\
                               & WaNet       & 83.47 & 81.28 & 95.66 & 98.54  \\
                               & Blend       & 74.01 & 71.02 & 87.37 & 94.96  \\
                               & Input-Aware & 15.89 &  3.99 & 23.77 & 62.91  \\
\midrule
\multirow{4}{*}{TinyImageNet} & BadNets     & 36.54 & 34.32 & 72.26 & 92.78  \\
                               & WaNet       & 53.18 & 50.94 & 82.14 & 95.76  \\
                               & Blend       & 89.84 & 89.54 & 94.76 & 97.52  \\
                               & Input-Aware & 10.88 &  9.52 & 28.76 & 64.68  \\
\bottomrule
\end{tabular}
\label{tab:unlearning_resnet18}
\end{table}

\begin{table}[!htb]
\centering
\small
\caption{FGA re-attack after trigger-aware unlearning on \textbf{VGG-19} across CIFAR-10, CIFAR-100, and TinyImageNet.}
\begin{tabular}{llccccc}
\toprule
Dataset & Attack & \shortstack{ASR after\\Unlearning} & \shortstack{FGA\\$\varepsilon{=}8/255$} & \shortstack{FGA\\$\varepsilon{=}16/255$} & \shortstack{FGA\\$\varepsilon{=}32/255$} \\
\midrule
\multirow{4}{*}{CIFAR-10}     & BadNets     & 97.42 & 97.37 & 99.65 & 99.86 \\
                               & WaNet       & 97.67 & 97.67 & 99.88 & 99.98 \\
                               & Blend       & 98.03 & 98.05 & 99.54 & 99.75 \\
                               & Input-Aware & 75.80 & 75.71 & 93.60 & 97.99 \\
\midrule
\multirow{4}{*}{CIFAR-100}    & BadNets     & 59.99 & 59.68 & 89.41 & 96.91 \\
                               & WaNet       & 76.59 & 76.52 & 96.11 & 99.61 \\
                               & Blend       & 74.32 & 74.27 & 88.46 & 95.00 \\
                               & Input-Aware & 27.41 & 24.86 & 73.16 & 95.82 \\
\midrule
\multirow{4}{*}{TinyImageNet} & BadNets     & 39.74 & 39.66 & 76.40 & 94.12 \\
                               & WaNet       & 32.10 & 31.98 & 64.84 & 90.54 \\
                               & Blend       & 96.08 & 96.06 & 97.92 & 99.18 \\
                               & Input-Aware & 11.38 & 11.50 & 36.42 & 67.06 \\
\bottomrule
\end{tabular}
\label{tab:unlearning_vgg19}
\end{table}

\section{Hyperparameter Analysis}
\label{sec:hyperparams}

\subsection{Effect of $\varepsilon$ (perturbation budget)}

\begin{figure}[htb]
    \centering
    \includegraphics[width=\linewidth]{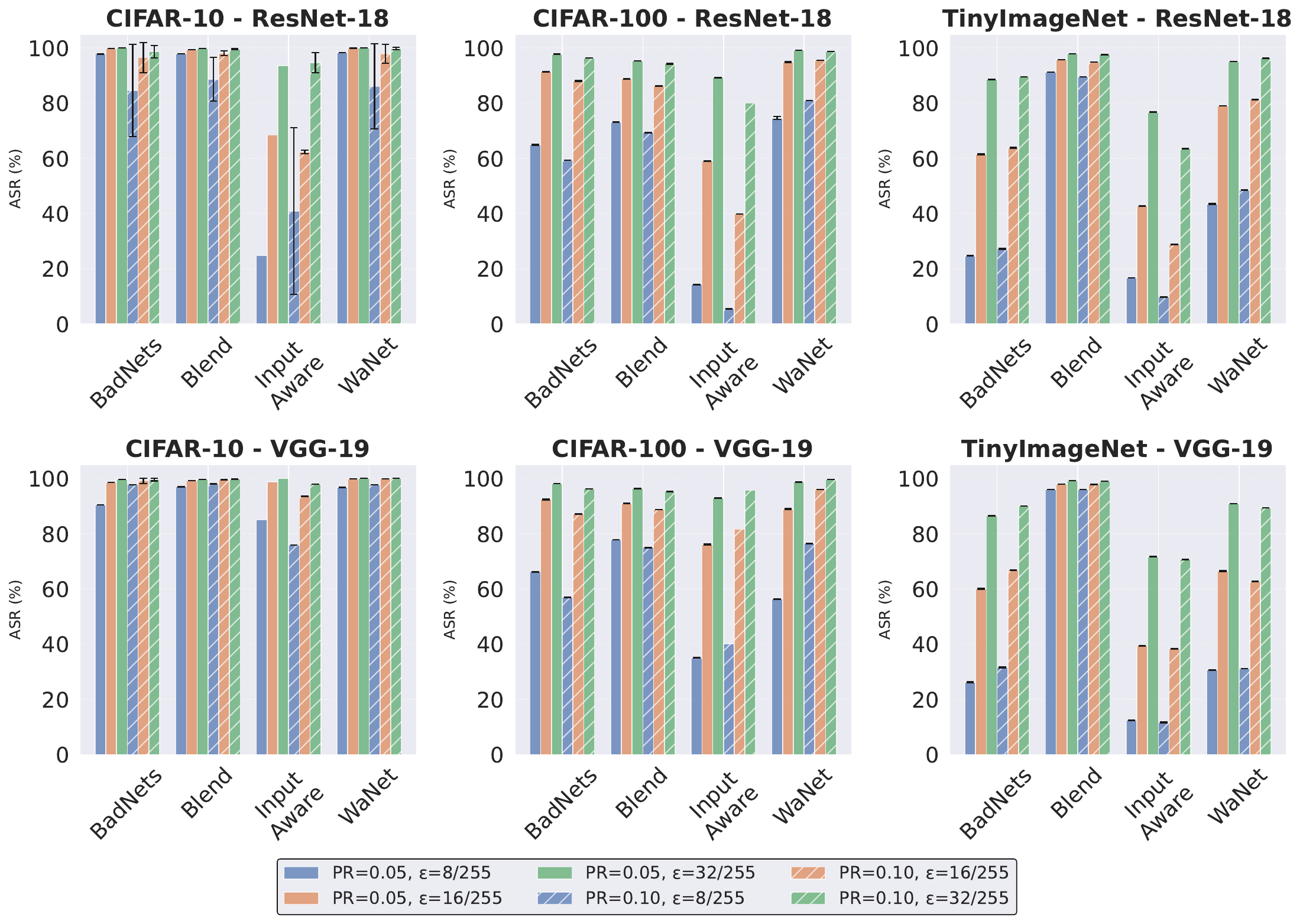}
    \caption{FGA success rate across perturbation budgets, poisoning rates (5\%, 10\%), 
    datasets, and architectures. Even at $\varepsilon=8/255$, most configurations exceed 85\% ASR.}
    \label{fig:fga_epsilon}
\end{figure}

Figure~\ref{fig:fga_epsilon} shows FGA performance across $\varepsilon \in \{8/255,\, 16/255,\, 32/255\}$, poisoning rates $\in \{0.05,\, 0.10\}$, and all dataset architecture combinations. As expected, larger perturbation budgets generally yield higher ASR. Importantly, even at $\varepsilon = 8/255$, FGA achieves high ASR across many settings, demonstrating that alternative triggers can be both subtle and effective. Input-Aware backdoors exhibit the weakest performance at small $\varepsilon$, likely due to their sample-specific trigger design, but still reach high ASR at larger budgets. Note that our goal is not to discover stealthy alternative triggers, so we do not enforce a limitation on the perturbation budget. Even though a small perturbation can also lead to high ASR.

\subsection{Effect of number of optimization steps.}
\label{sec:optimization_steps}

\begin{figure*}[htb]
  \centering
  \begin{subfigure}{0.32\textwidth}
    \centering
    \includegraphics[width=\linewidth]{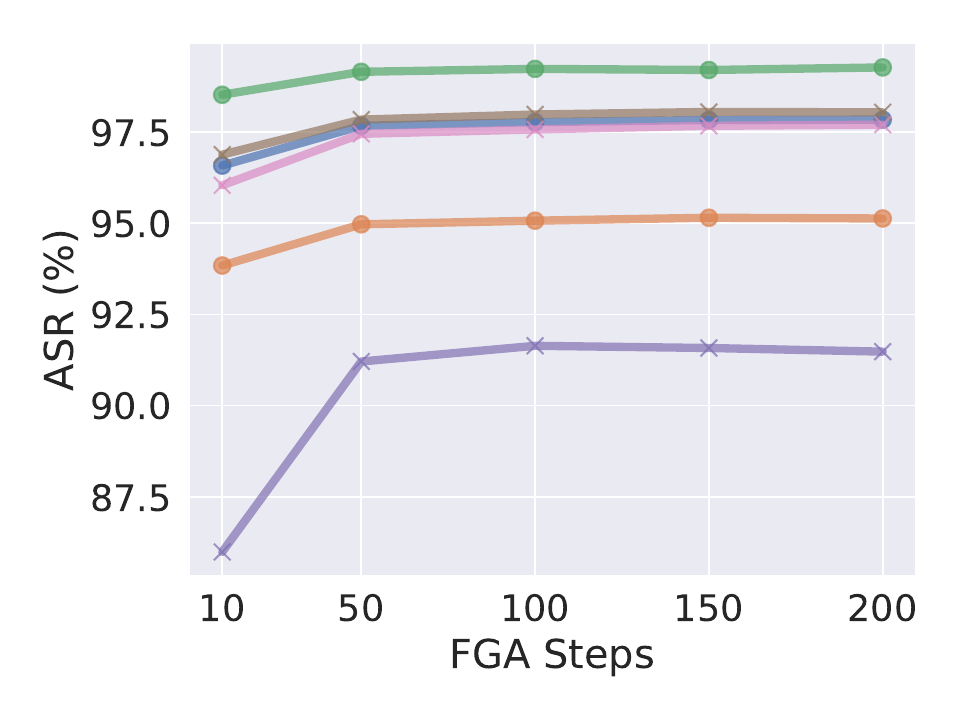}
    \caption{$\varepsilon = 8 / 255$}
  \end{subfigure}
  \hfill
  \begin{subfigure}{0.32\textwidth}
    \centering
    \includegraphics[width=\linewidth]{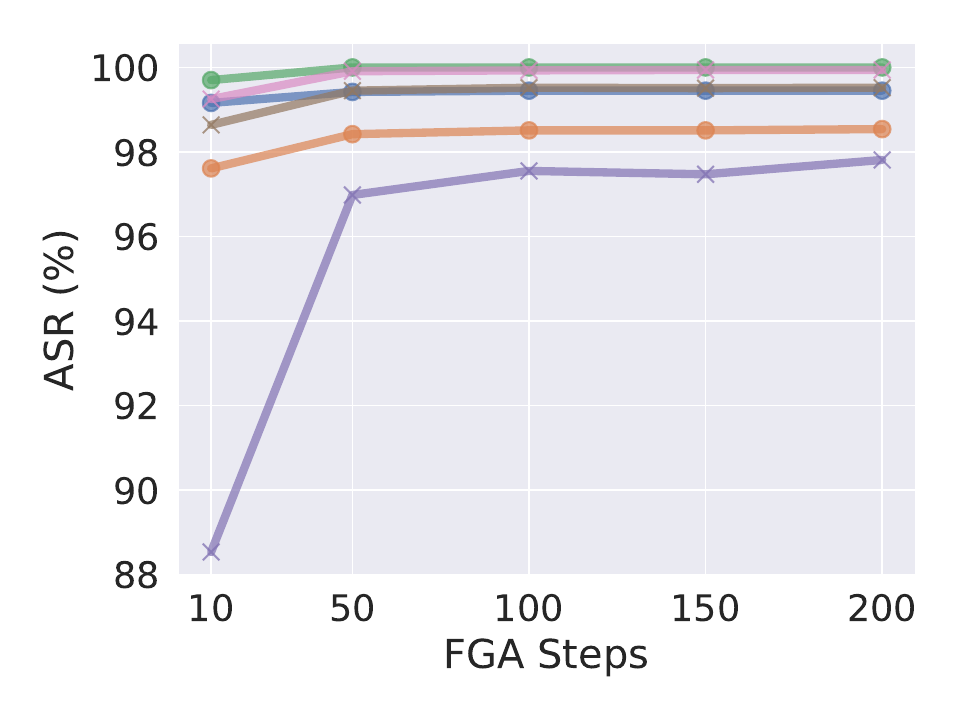}
    \caption{$\varepsilon = 16 / 255$}
  \end{subfigure}
  \hfill
    \begin{subfigure}{0.32\textwidth}
    \centering
    \includegraphics[width=\linewidth]{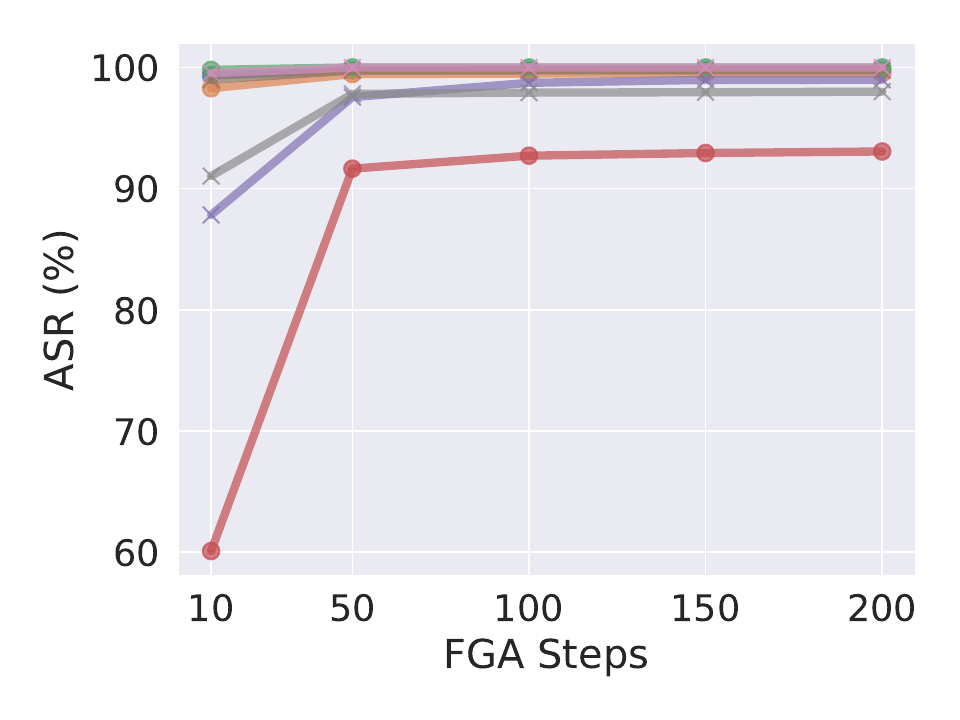}
    \caption{ $\varepsilon = 32 / 255$}
  \end{subfigure}
  \caption{Effect of optimization steps for CIFAR-10 under different configurations. Line colors denote model and attack:
{\color{blue}$\blacksquare$} ResNet-18 BadNets,
{\color{orange}$\blacksquare$} ResNet-18 Blend,
{\color{green}$\blacksquare$} ResNet-18 WaNet,
{\color{red}$\blacksquare$} ResNet-18 InputAware,
{\color{violet}$\blacksquare$} VGG-19 BadNets,
{\color{brown}$\blacksquare$} VGG-19 Blend,
{\color{pink}$\blacksquare$} VGG-19 WaNet,
{\color{darkgray}$\blacksquare$} VGG-19 InputAware.}
\label{fig:steps_cifar10}
\end{figure*}

\begin{figure*}[htb]
  \centering
  \begin{subfigure}{0.32\textwidth}
    \centering
    \includegraphics[width=\linewidth]{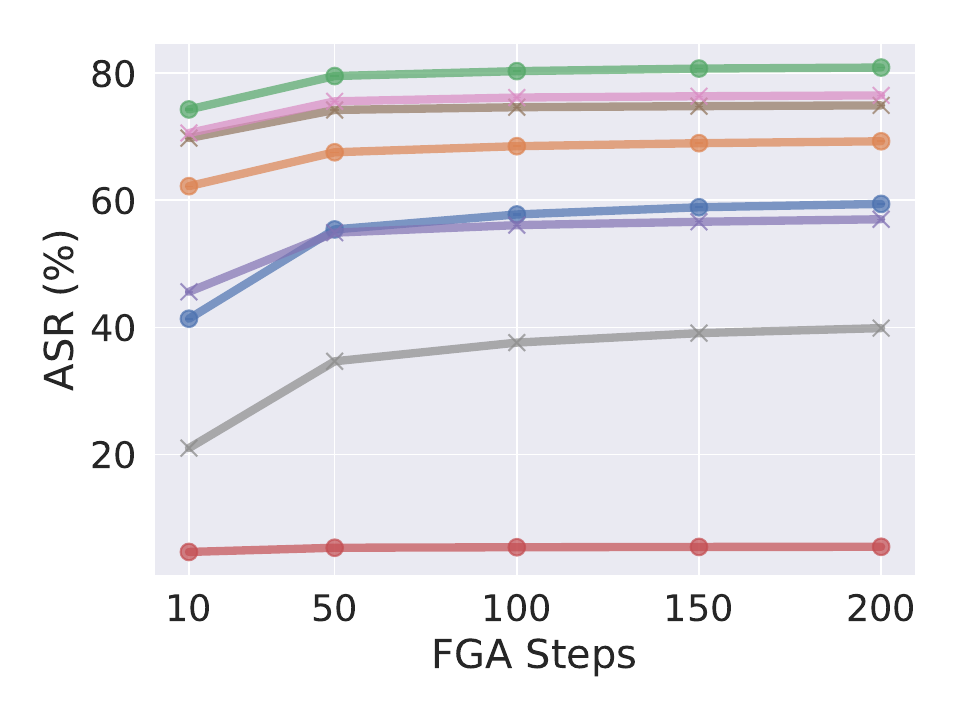}
    \caption{$\varepsilon = 8 / 255$}
  \end{subfigure}
  \hfill
  \begin{subfigure}{0.32\textwidth}
    \centering
    \includegraphics[width=\linewidth]{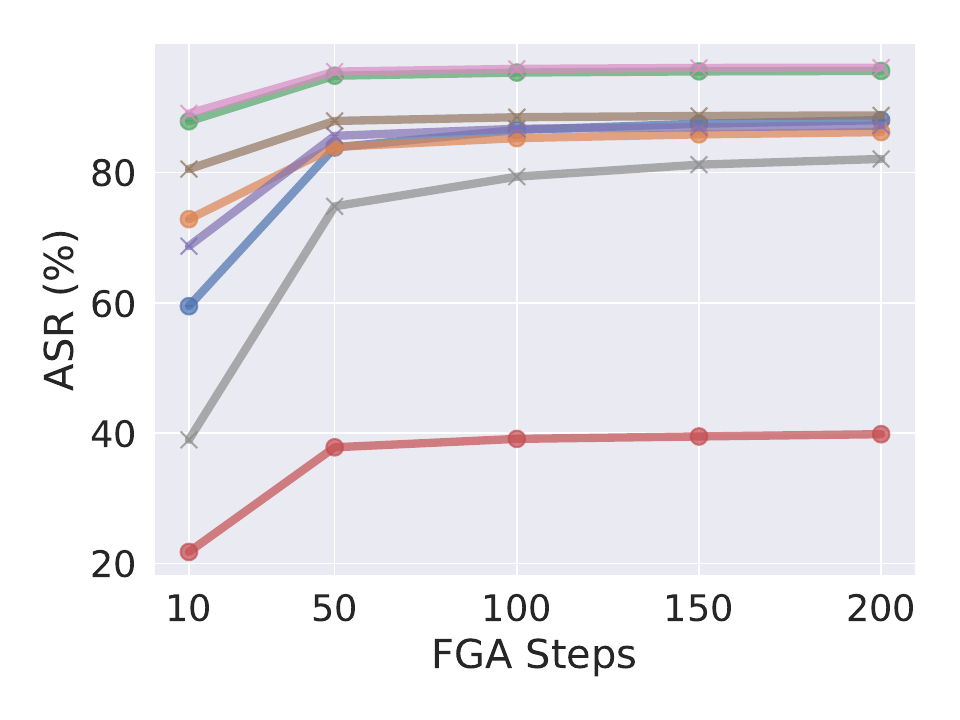}
    \caption{$\varepsilon = 16 / 255$}
  \end{subfigure}
  \hfill
    \begin{subfigure}{0.32\textwidth}
    \centering
    \includegraphics[width=\linewidth]{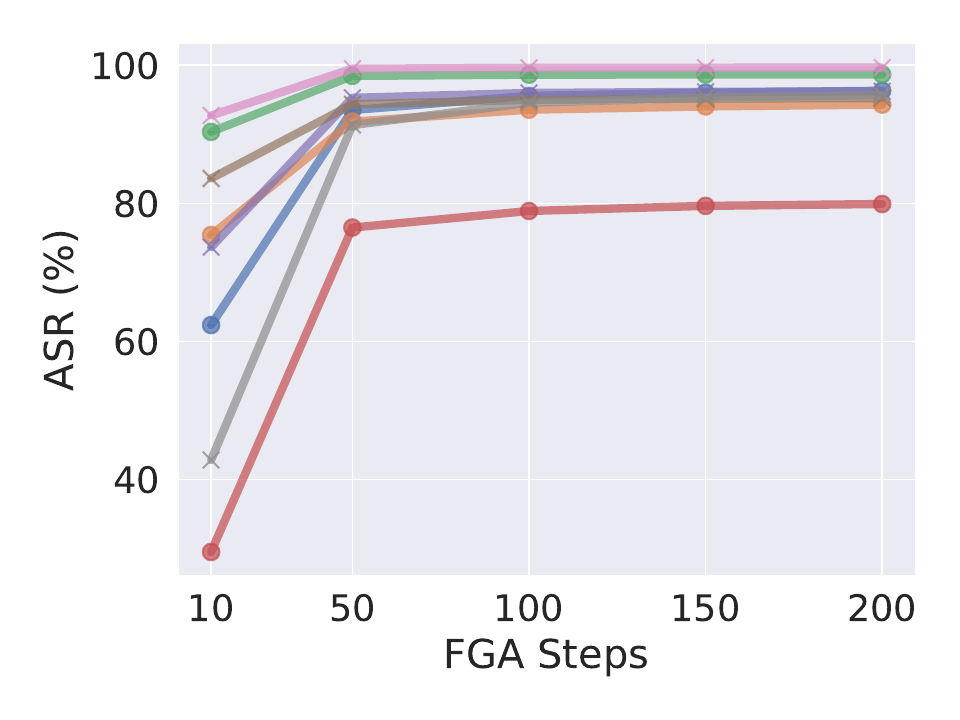}
    \caption{$\varepsilon = 32 / 255$}
  \end{subfigure}
  \caption{Effect of optimization steps for CIFAR-100 under different configurations. Line colors denote model and attack:
{\color{blue}$\blacksquare$} ResNet-18 BadNets,
{\color{orange}$\blacksquare$} ResNet-18 Blend,
{\color{green}$\blacksquare$} ResNet-18 WaNet,
{\color{red}$\blacksquare$} ResNet-18 InputAware,
{\color{violet}$\blacksquare$} VGG-19 BadNets,
{\color{brown}$\blacksquare$} VGG-19 Blend,
{\color{pink}$\blacksquare$} VGG-19 WaNet,
{\color{darkgray}$\blacksquare$} VGG-19 InputAware.}
    \label{fig:steps_cifar100}
\end{figure*}

\begin{figure*}[htb]
  \centering
  \begin{subfigure}{0.32\textwidth}
    \centering
    \includegraphics[width=\linewidth]{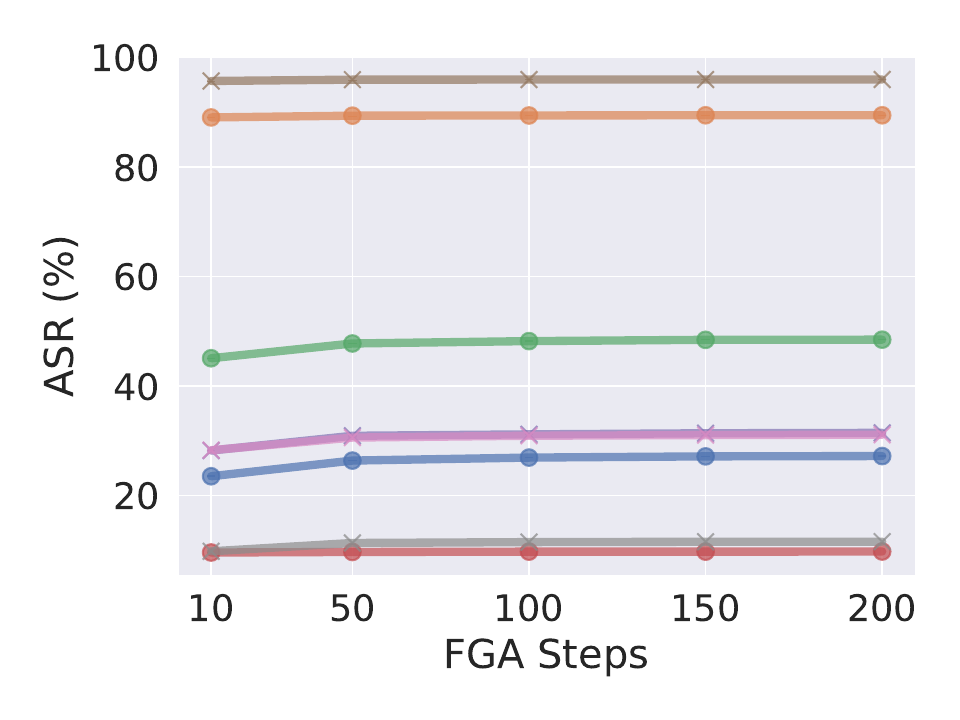}
    \caption{$\varepsilon = 8 / 255$}
  \end{subfigure}
  \hfill
  \begin{subfigure}{0.32\textwidth}
    \centering
    \includegraphics[width=\linewidth]{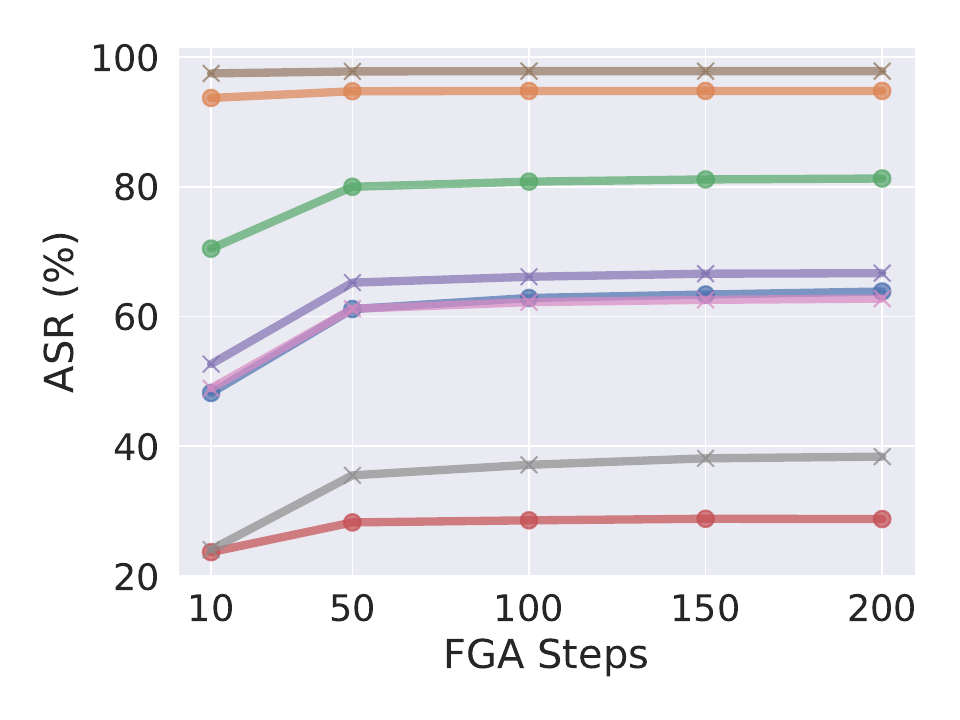}
    \caption{$\varepsilon = 16 / 255$}
  \end{subfigure}
  \hfill
    \begin{subfigure}{0.32\textwidth}
    \centering
    \includegraphics[width=\linewidth]{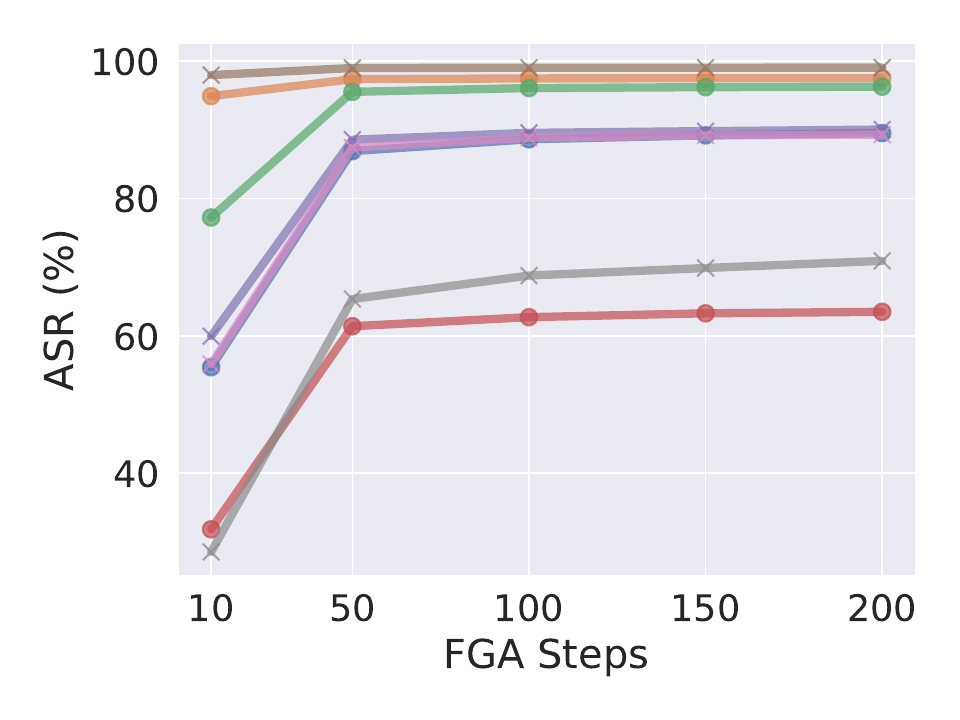}
    \caption{$\varepsilon = 32 / 255$}
  \end{subfigure}
  \caption{Effect of optimization steps for TinyImageNet under different configurations. Line colors denote model and attack:
{\color{blue}$\blacksquare$} ResNet-18 BadNets,
{\color{orange}$\blacksquare$} ResNet-18 Blend,
{\color{green}$\blacksquare$} ResNet-18 WaNet,
{\color{red}$\blacksquare$} ResNet-18 InputAware,
{\color{violet}$\blacksquare$} VGG-19 BadNets,
{\color{brown}$\blacksquare$} VGG-19 Blend,
{\color{pink}$\blacksquare$} VGG-19 WaNet,
{\color{darkgray}$\blacksquare$} VGG-19 InputAware.}
\label{fig:steps_tiny} 
  
\end{figure*}

Figures~\ref{fig:steps_cifar10},~\ref{fig:steps_cifar100}, and~\ref{fig:steps_tiny} shows ASR as a function of FGA optimization steps on CIFAR-10, CIFAR-100, and TinyImagNet respectively. Most configurations reach above 80\% ASR within 50 steps, with marginal gains beyond 100 steps. This indicates that the estimated backdoor direction $d_\ell$ provides a strong optimization signal that quickly guides perturbations into the backdoor region. Note that TinyImageNet is more complicated to achieve a high ASR due to having more classes and thus a more complex latent space. This is noticeable with CIFAR-100. However, under large perturbation budgets, the ASR rapidly increases. Also note that the model architecture plays an important role. 

\subsection{Backdoor Alignment and Attack Success.}
\label{sec:backdoor_aligment}

\begin{figure*}[htb]
  \centering

  \begin{subfigure}{0.32\textwidth}
    \centering
    \includegraphics[width=\linewidth]{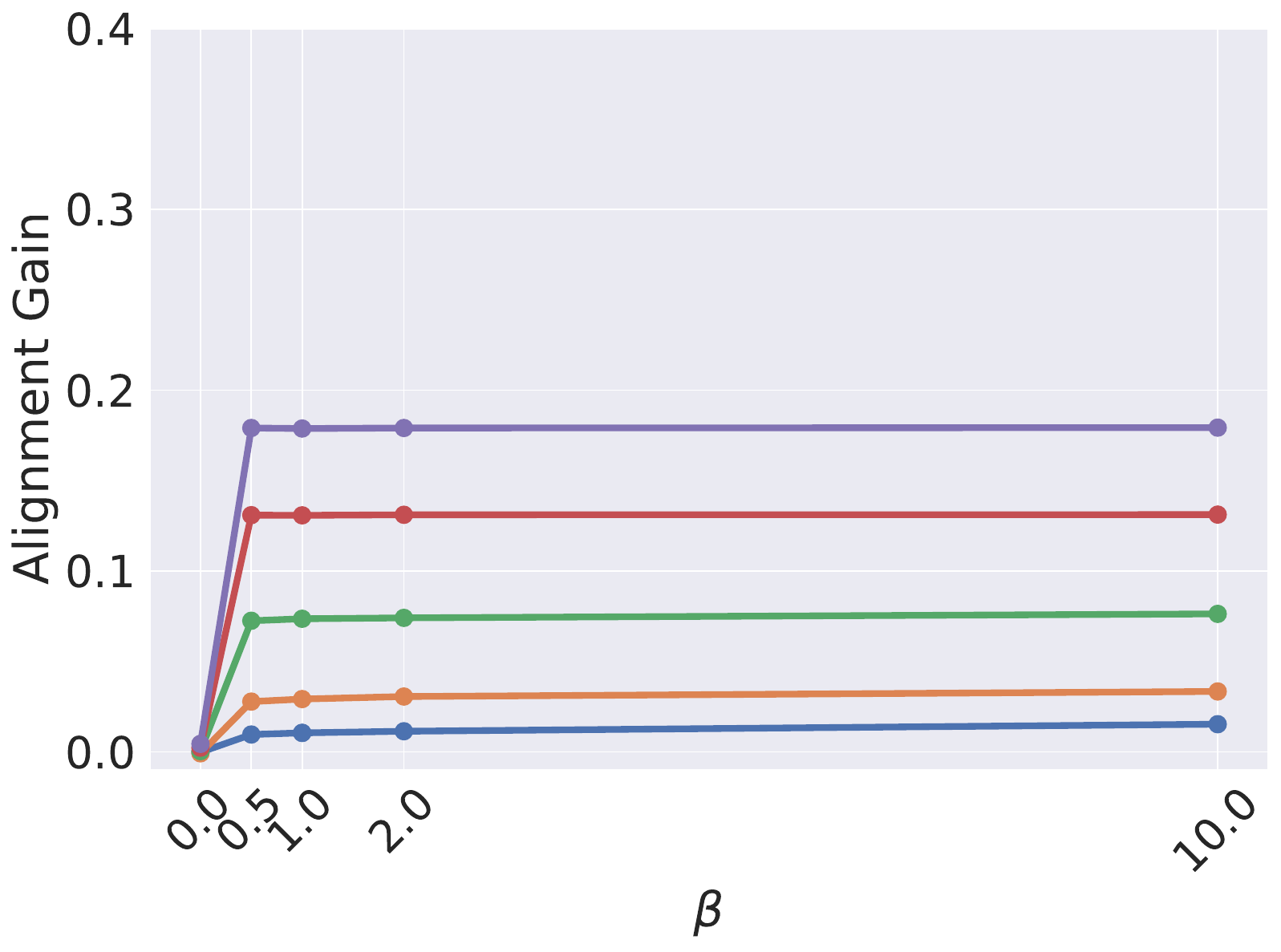}
    \caption{ResNet-18, CIFAR-10, BadNets}
  \end{subfigure}
  \hfill
  \begin{subfigure}{0.32\textwidth}
    \centering
    \includegraphics[width=\linewidth]{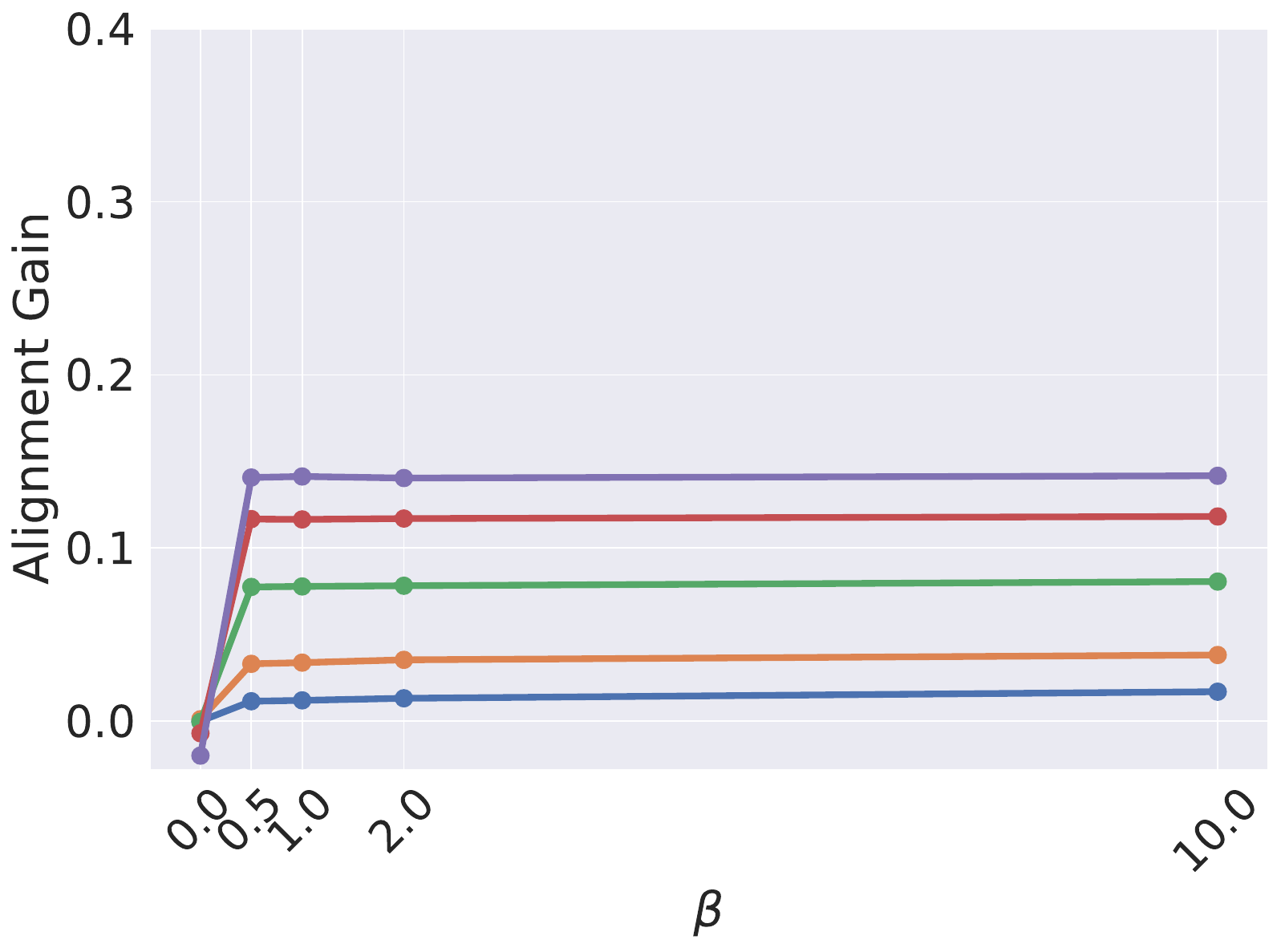}
    \caption{ResNet-18, CIFAR-10, Blend}
  \end{subfigure}
  \hfill
  \begin{subfigure}{0.32\textwidth}
    \centering
    \includegraphics[width=\linewidth]{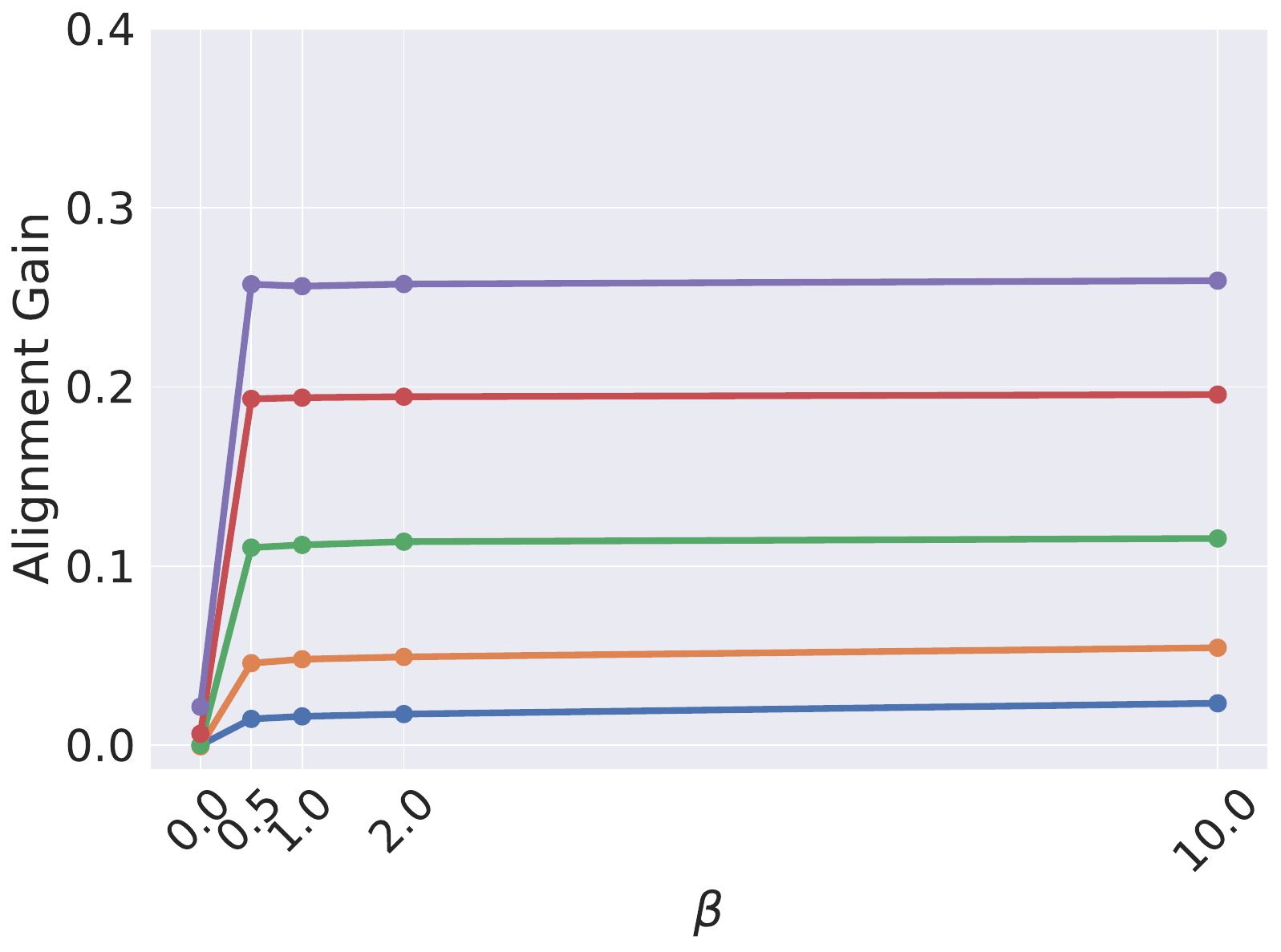}
    \caption{ResNet-18, CIFAR-10, WaNet}
  \end{subfigure}

  \medskip

  \begin{subfigure}{0.32\textwidth}
    \centering
    \includegraphics[width=\linewidth]{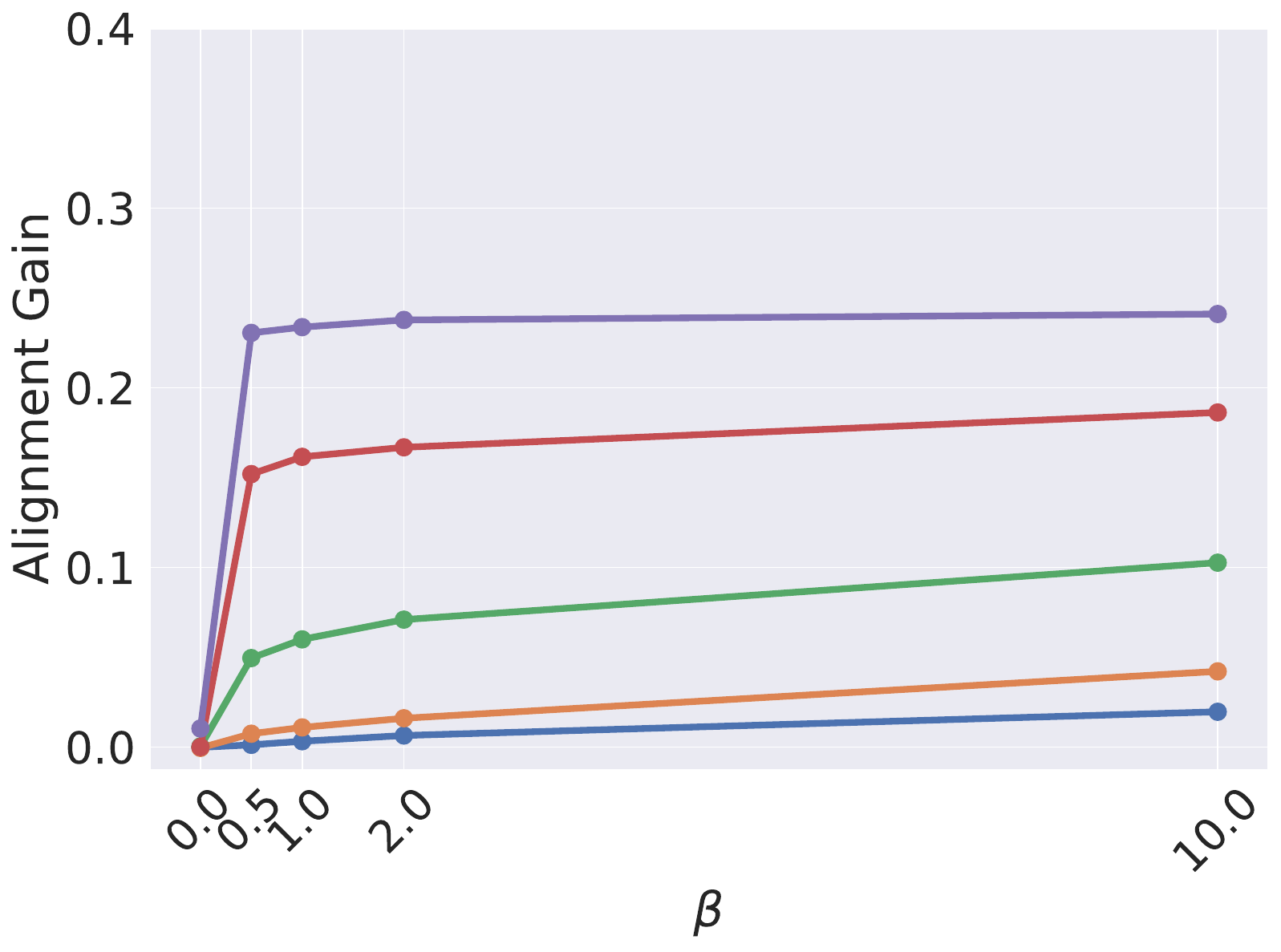}
    \caption{ResNet-18, CIFAR-100, BadNets}
  \end{subfigure}
  \hfill
  \begin{subfigure}{0.32\textwidth}
    \centering
    \includegraphics[width=\linewidth]{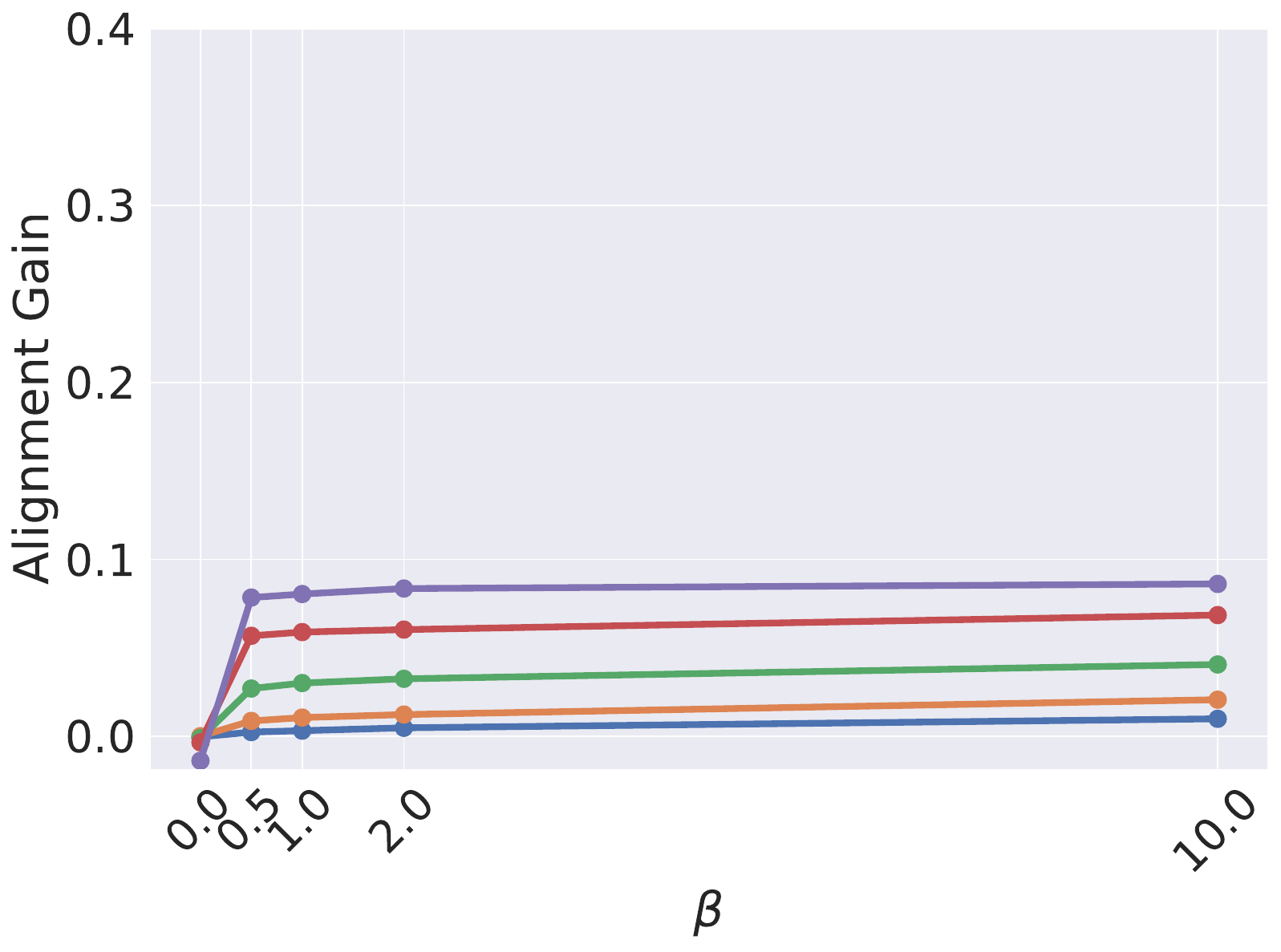}
    \caption{ResNet-18, CIFAR-100, Blend}
  \end{subfigure}
  \hfill
  \begin{subfigure}{0.32\textwidth}
    \centering
    \includegraphics[width=\linewidth]{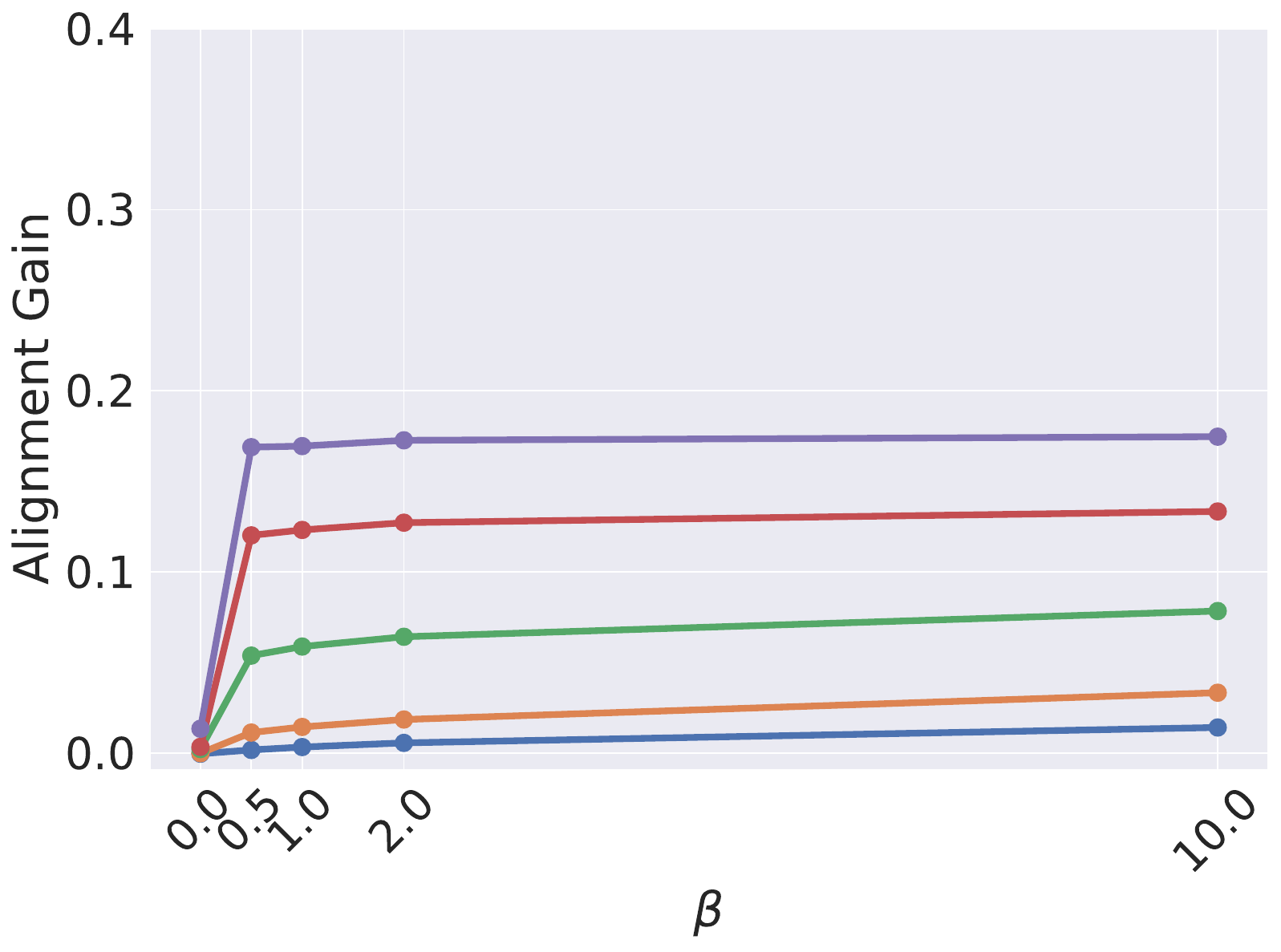}
    \caption{ResNet-18, CIFAR-100, WaNet}
  \end{subfigure}

  \medskip

  \begin{subfigure}{0.32\textwidth}
    \centering
    \includegraphics[width=\linewidth]{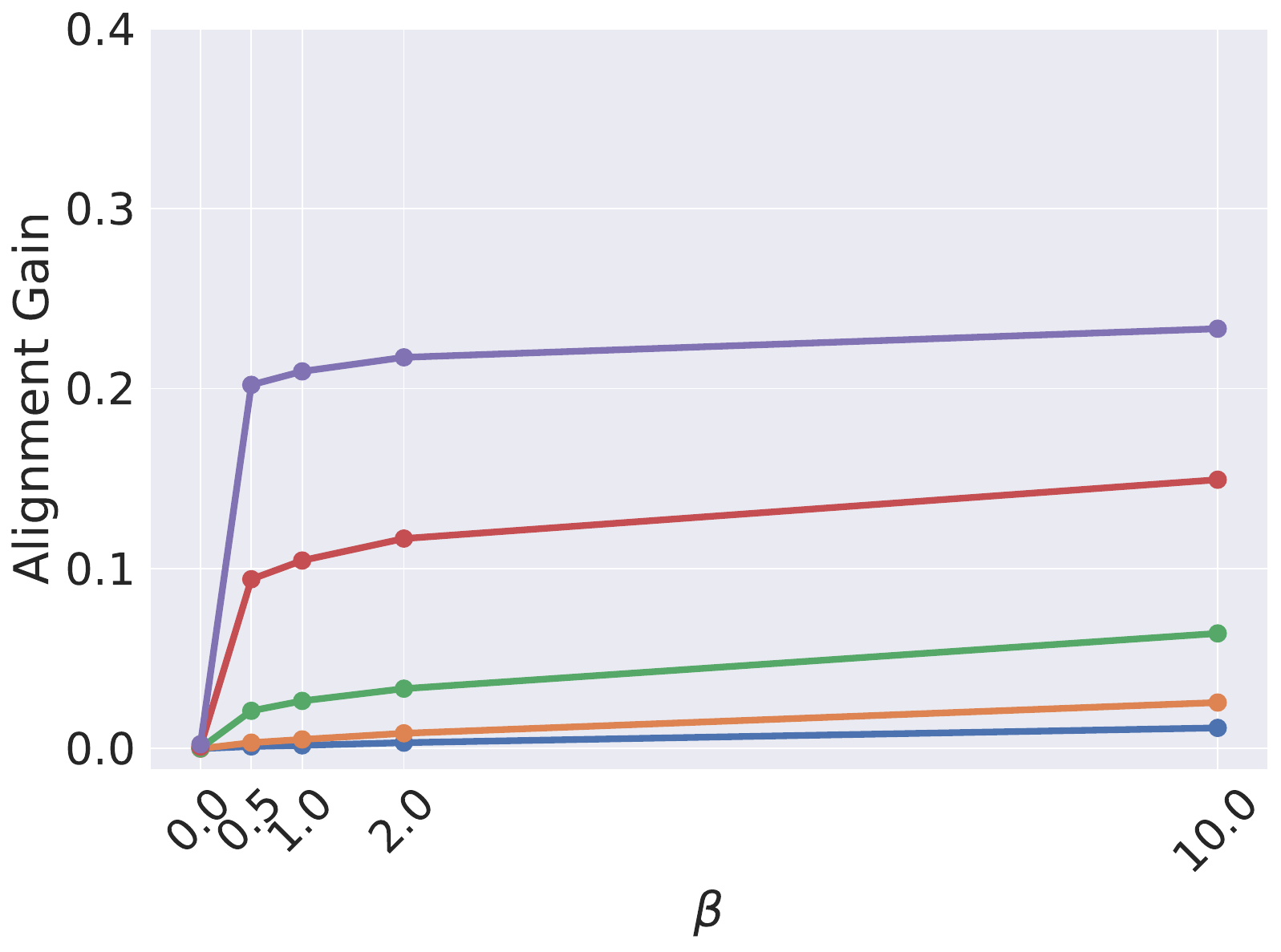}
    \caption{ResNet-18, TinyImageNet, BadNets}
  \end{subfigure}
  \hfill
  \begin{subfigure}{0.32\textwidth}
    \centering
    \includegraphics[width=\linewidth]{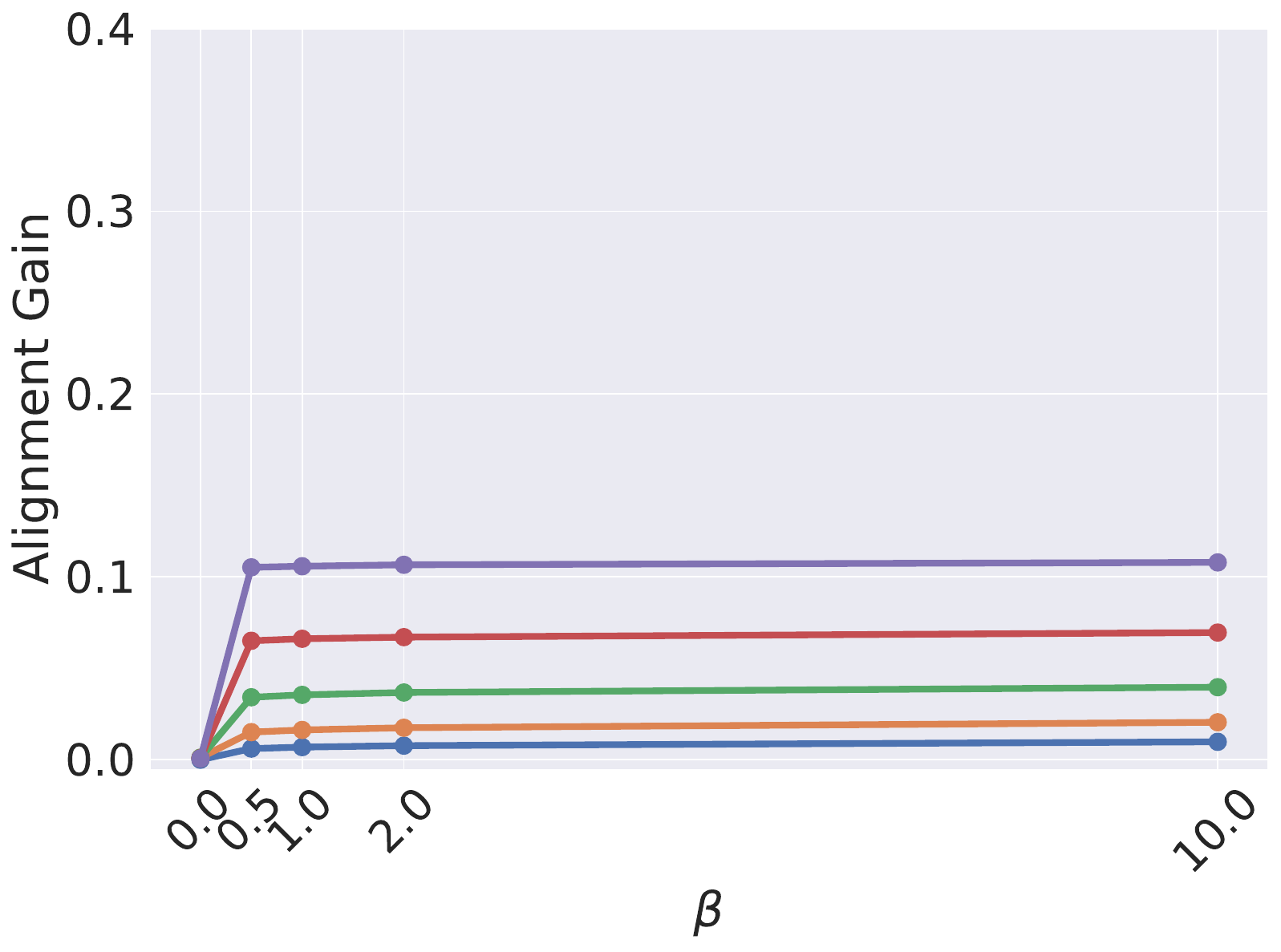}
    \caption{ResNet-18, TinyImageNet, Blend}
  \end{subfigure}
  \hfill
  \begin{subfigure}{0.32\textwidth}
    \centering
    \includegraphics[width=\linewidth]{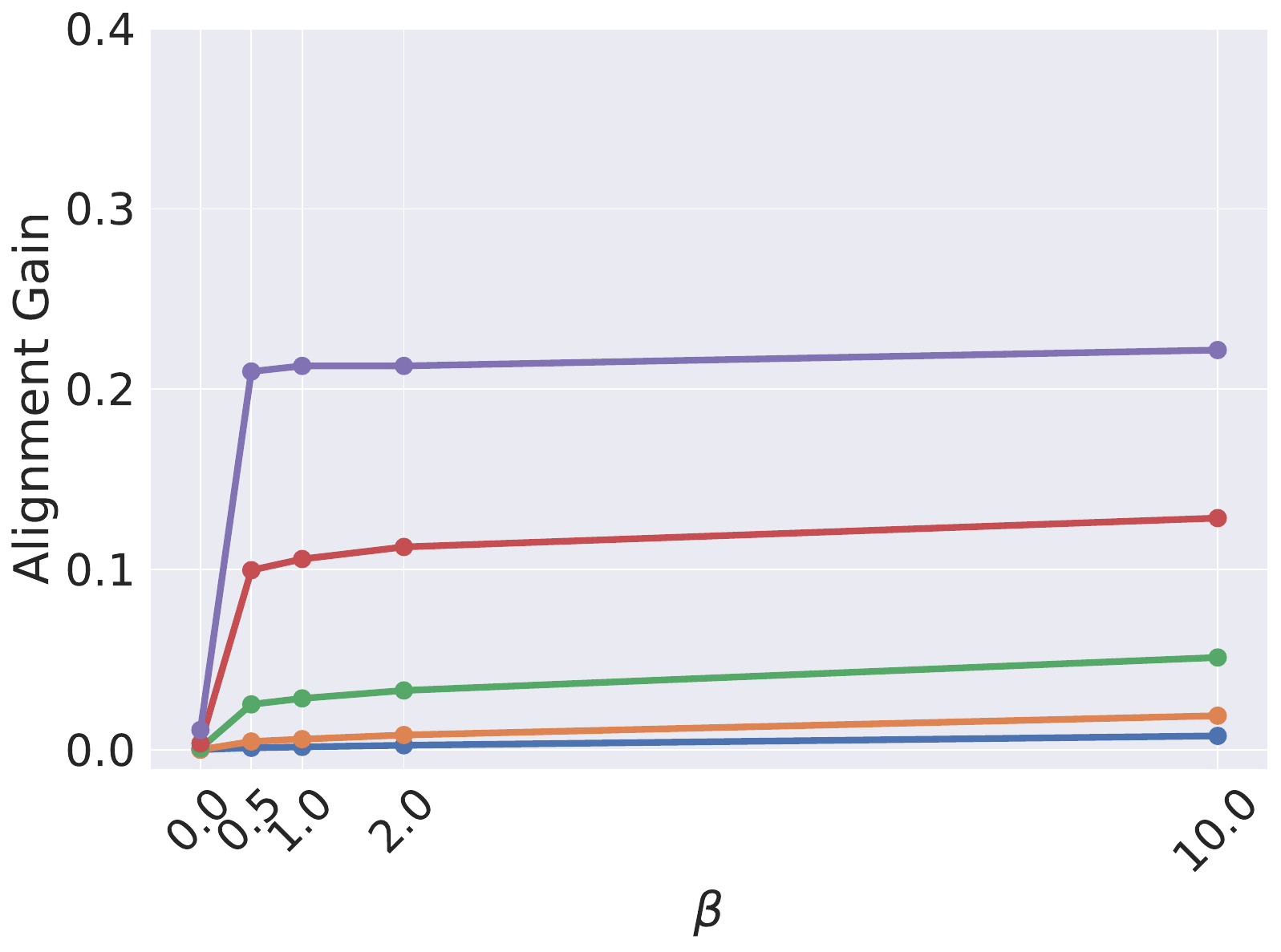}
    \caption{ResNet-18, TinyImageNet, WaNet}
  \end{subfigure}

  \medskip

  \begin{subfigure}{0.32\textwidth}
    \centering
    \includegraphics[width=\linewidth]{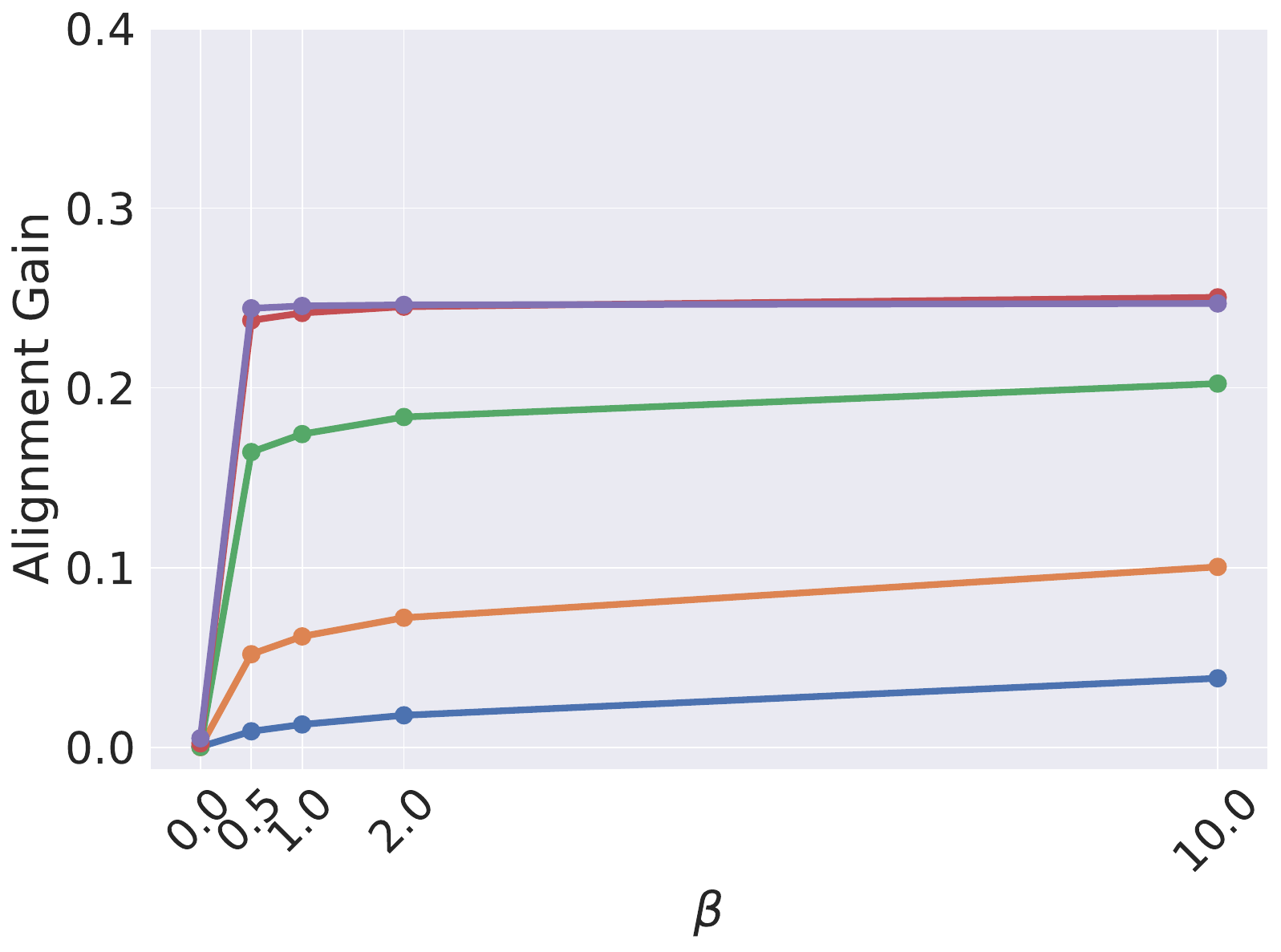}
    \caption{ResNet-18, TinyImageNet-224, BadNets}
  \end{subfigure}
  \hfill
  \begin{subfigure}{0.32\textwidth}
    \centering
    \includegraphics[width=\linewidth]{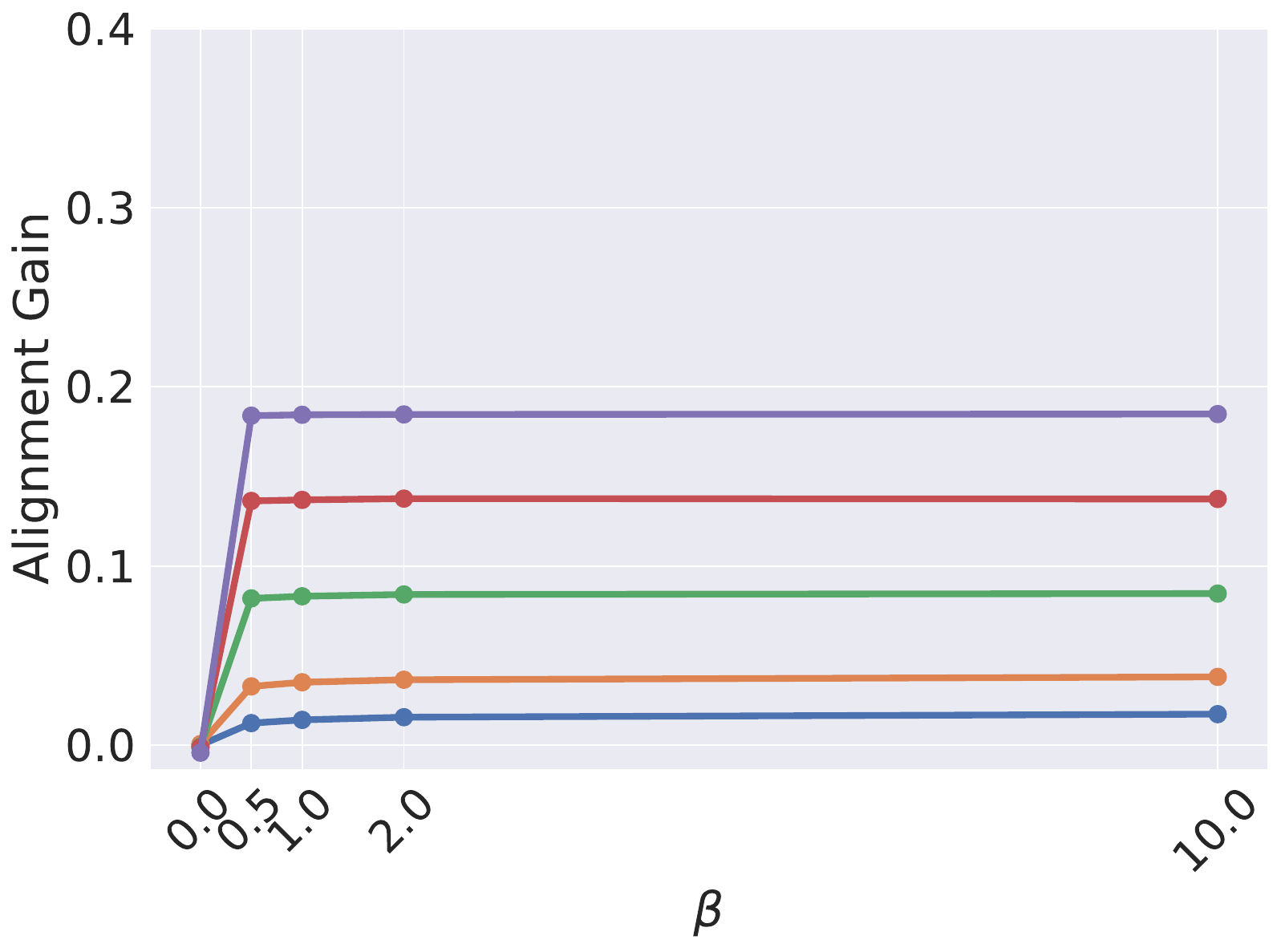}
    \caption{ResNet-18, TinyImageNet-224, Blend}
  \end{subfigure}
  \hfill
  \begin{subfigure}{0.32\textwidth}
    \centering
    \includegraphics[width=\linewidth]{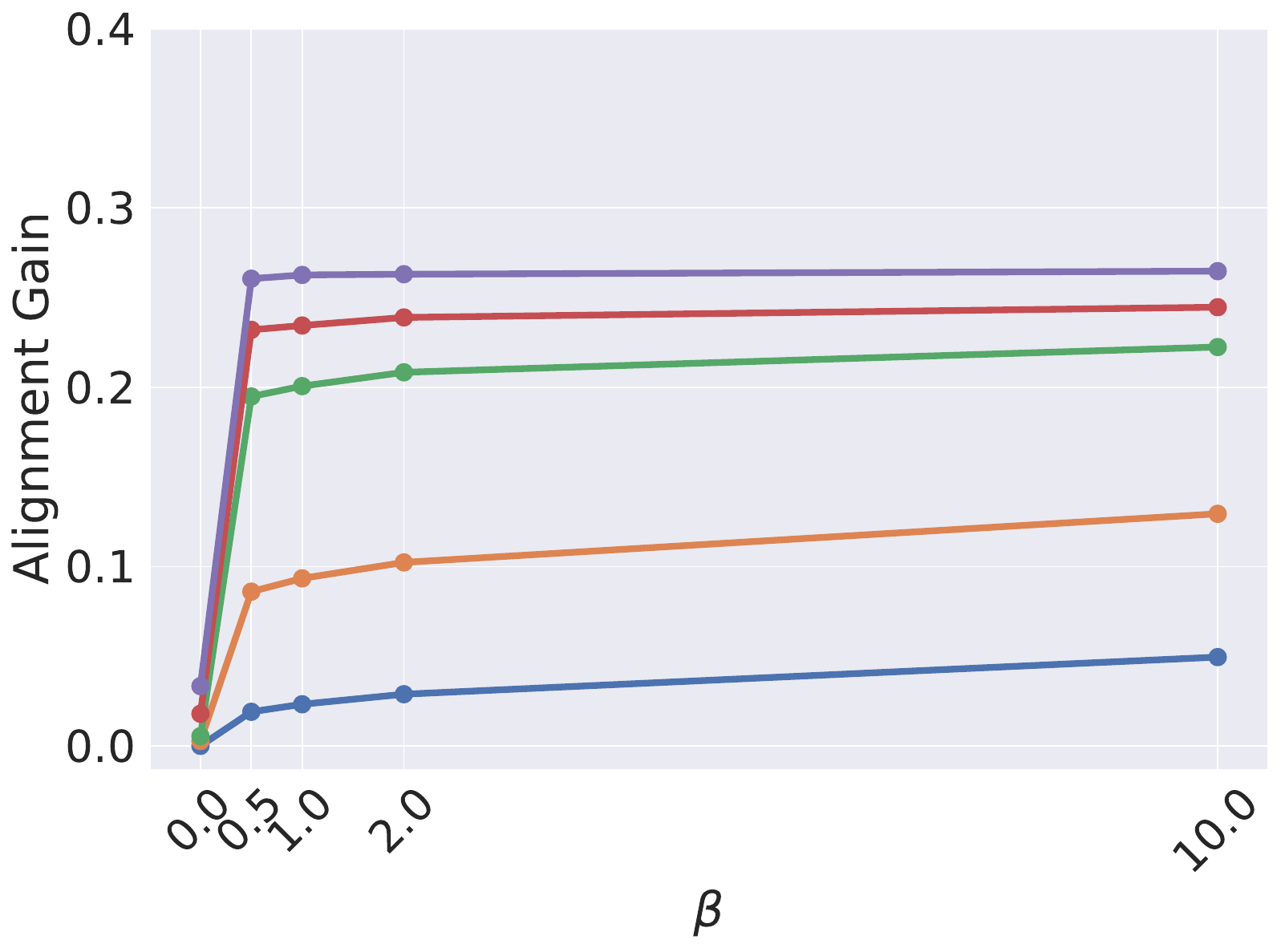}
    \caption{ResNet-18, TinyImageNet-224, WaNet}
  \end{subfigure}

  \caption{Trade-off between alignment and performance for ResNet-18 across datasets (CIFAR-10, CIFAR-100, TinyImageNet, TinyImageNet-224) and attacks (BadNets, Blend, WaNet). Blue = 2/255, orange = 4/255, green = 8/255, pink = 16/266, and purple = 32/255.}
  \label{fig:resnet18_alignment_tradeoff}
\end{figure*}

\begin{figure*}[htb]
  \centering

  \begin{subfigure}{0.32\textwidth}
    \centering
    \includegraphics[width=\linewidth]{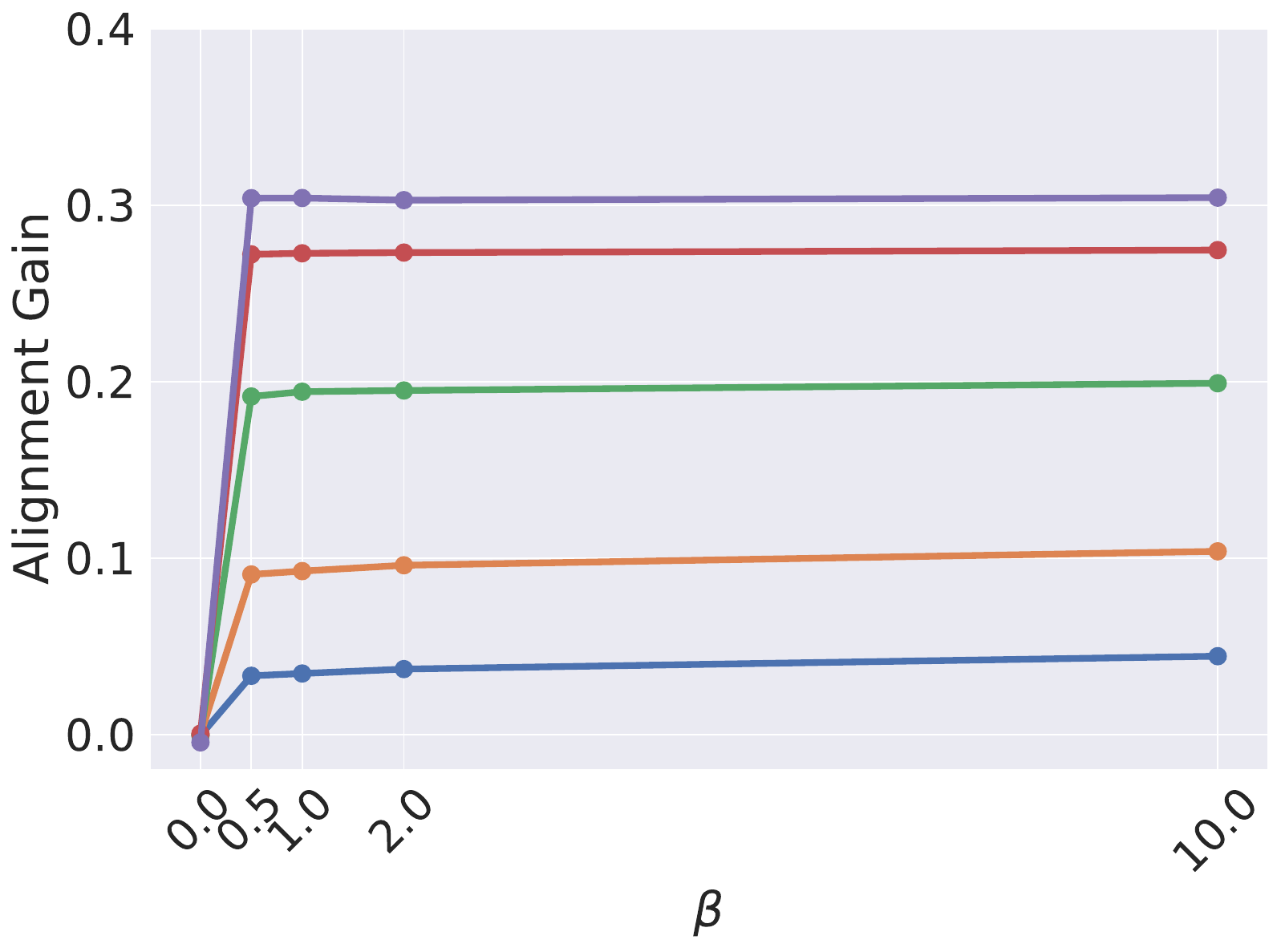}
    \caption{VGG-19, CIFAR-10, BadNets}
  \end{subfigure}
  \hfill
  \begin{subfigure}{0.32\textwidth}
    \centering
    \includegraphics[width=\linewidth]{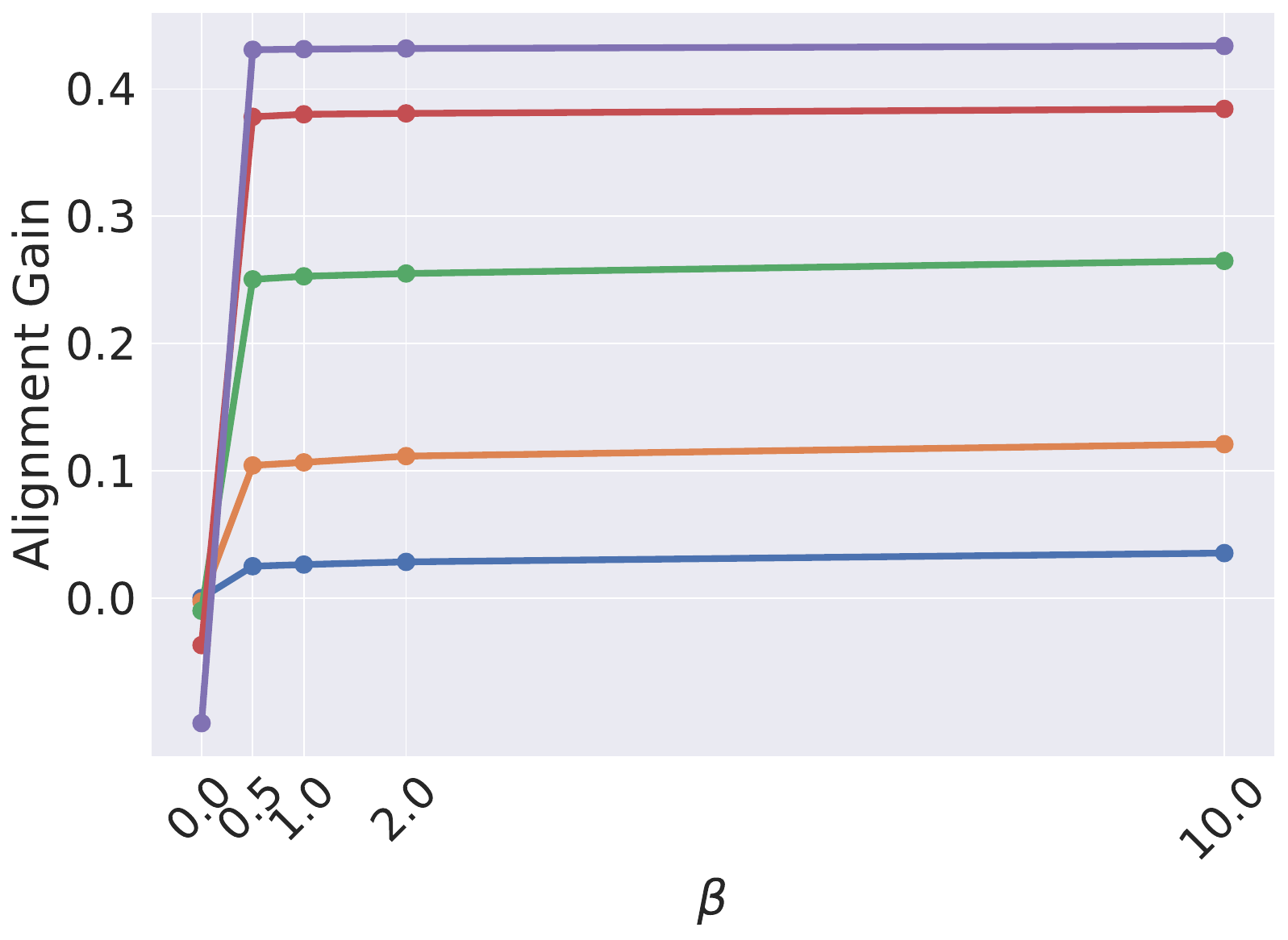}
    \caption{VGG-19, CIFAR-10, Blend}
  \end{subfigure}
  \hfill
  \begin{subfigure}{0.32\textwidth}
    \centering
    \includegraphics[width=\linewidth]{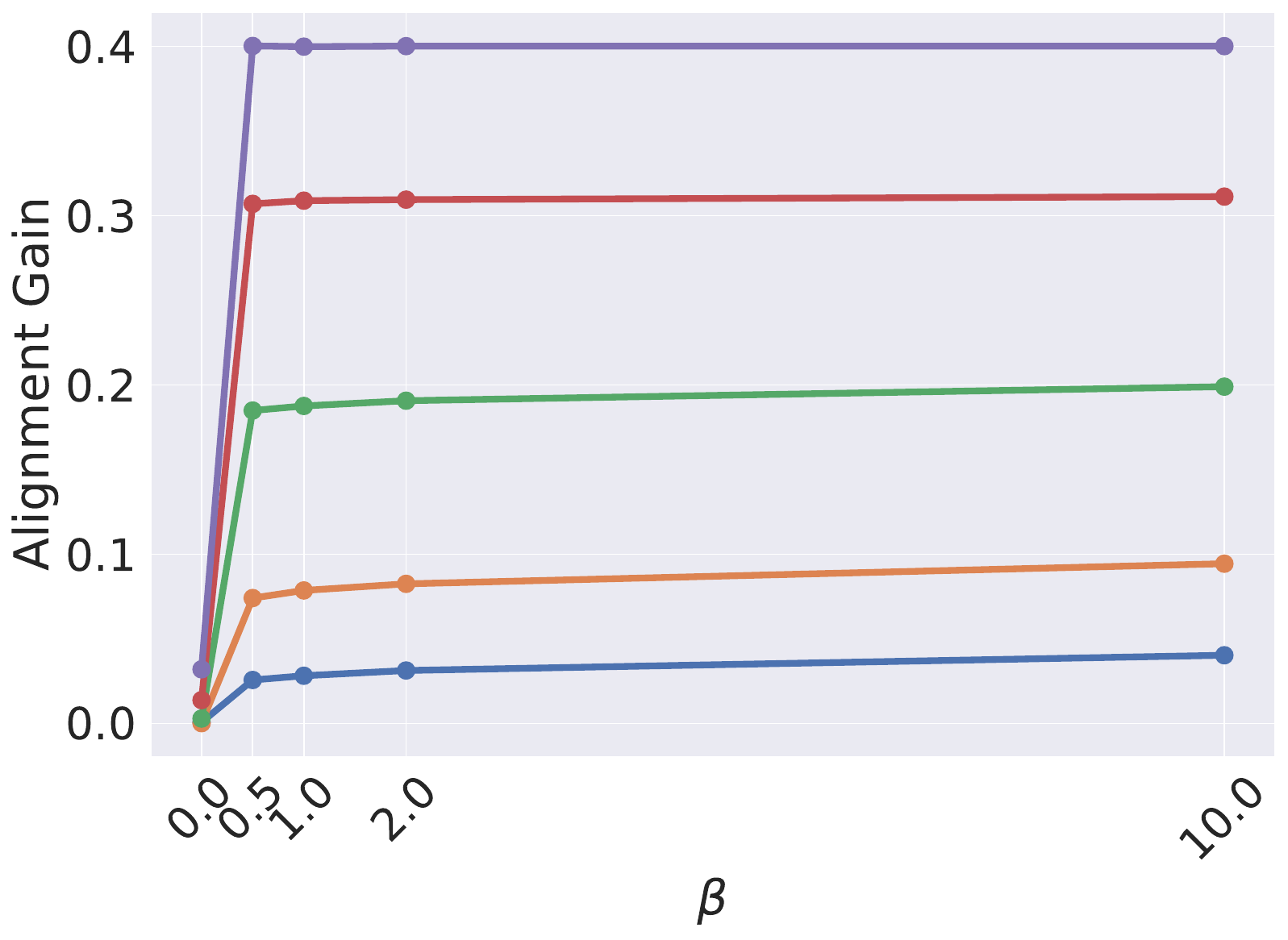}
    \caption{VGG-19, CIFAR-10, WaNet}
  \end{subfigure}

  \medskip

  \begin{subfigure}{0.32\textwidth}
    \centering
    \includegraphics[width=\linewidth]{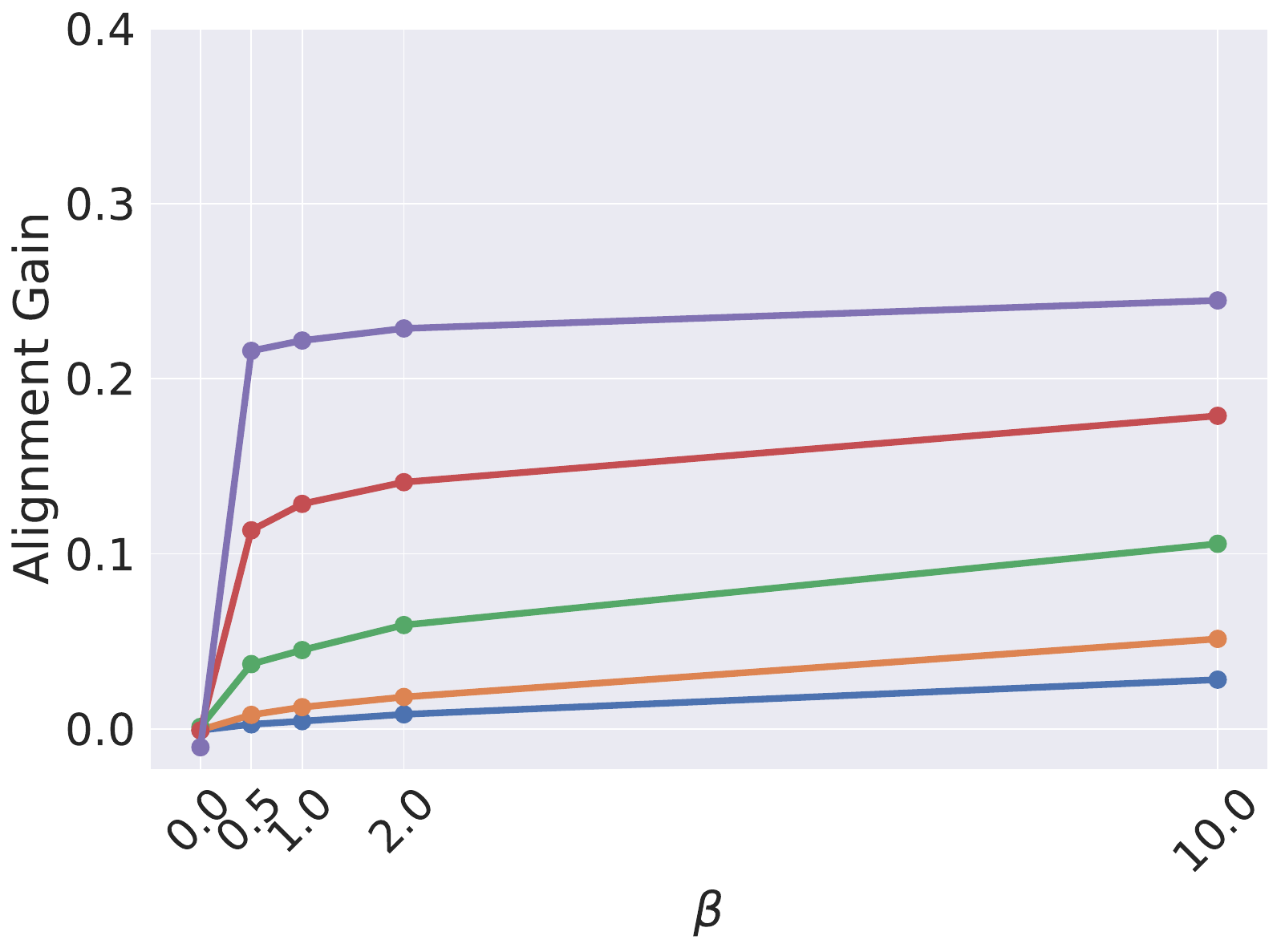}
    \caption{VGG-19, CIFAR-100, BadNets}
  \end{subfigure}
  \hfill
  \begin{subfigure}{0.32\textwidth}
    \centering
    \includegraphics[width=\linewidth]{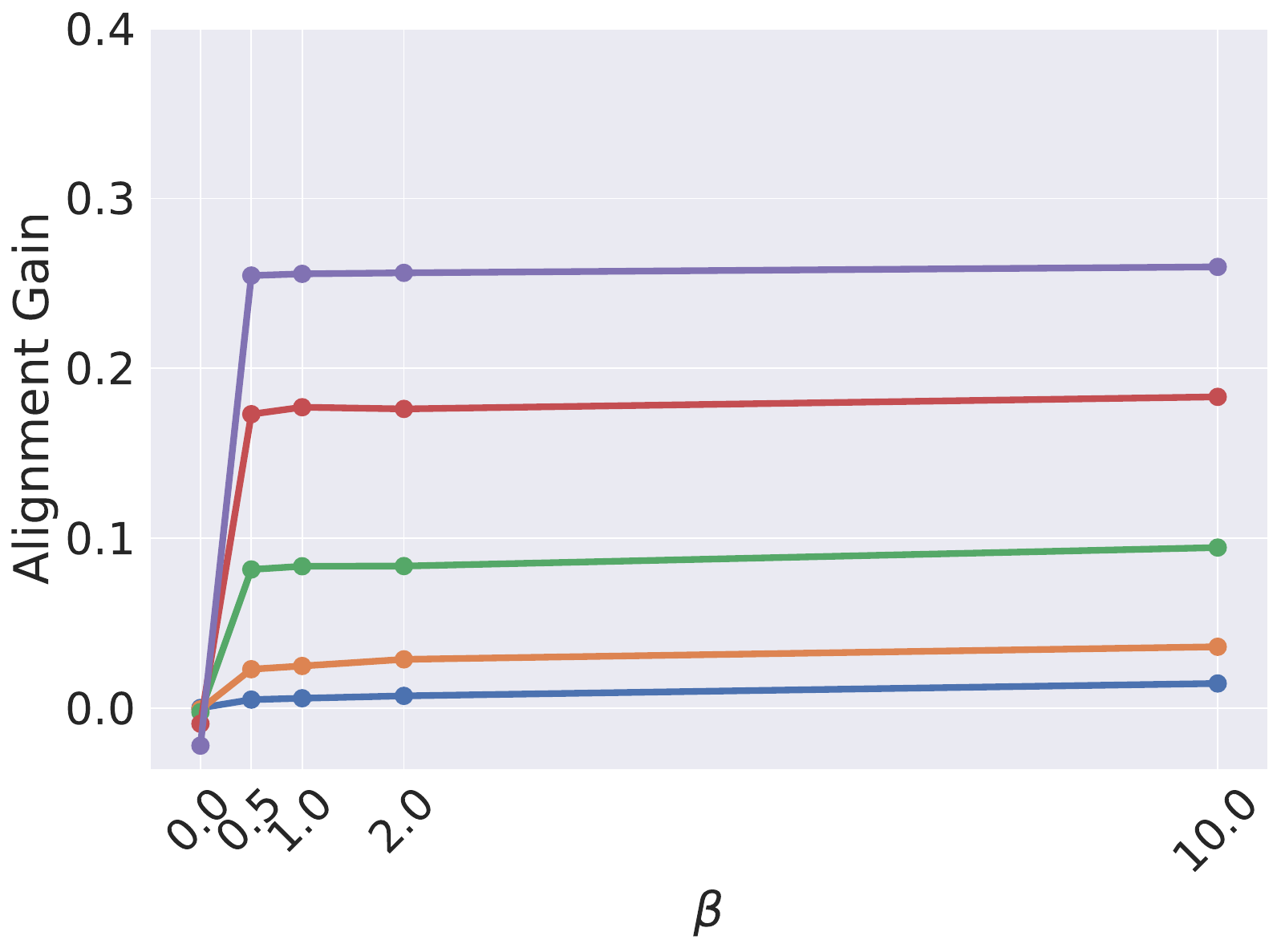}
    \caption{VGG-19, CIFAR-100, Blend}
  \end{subfigure}
  \hfill
  \begin{subfigure}{0.32\textwidth}
    \centering
    \includegraphics[width=\linewidth]{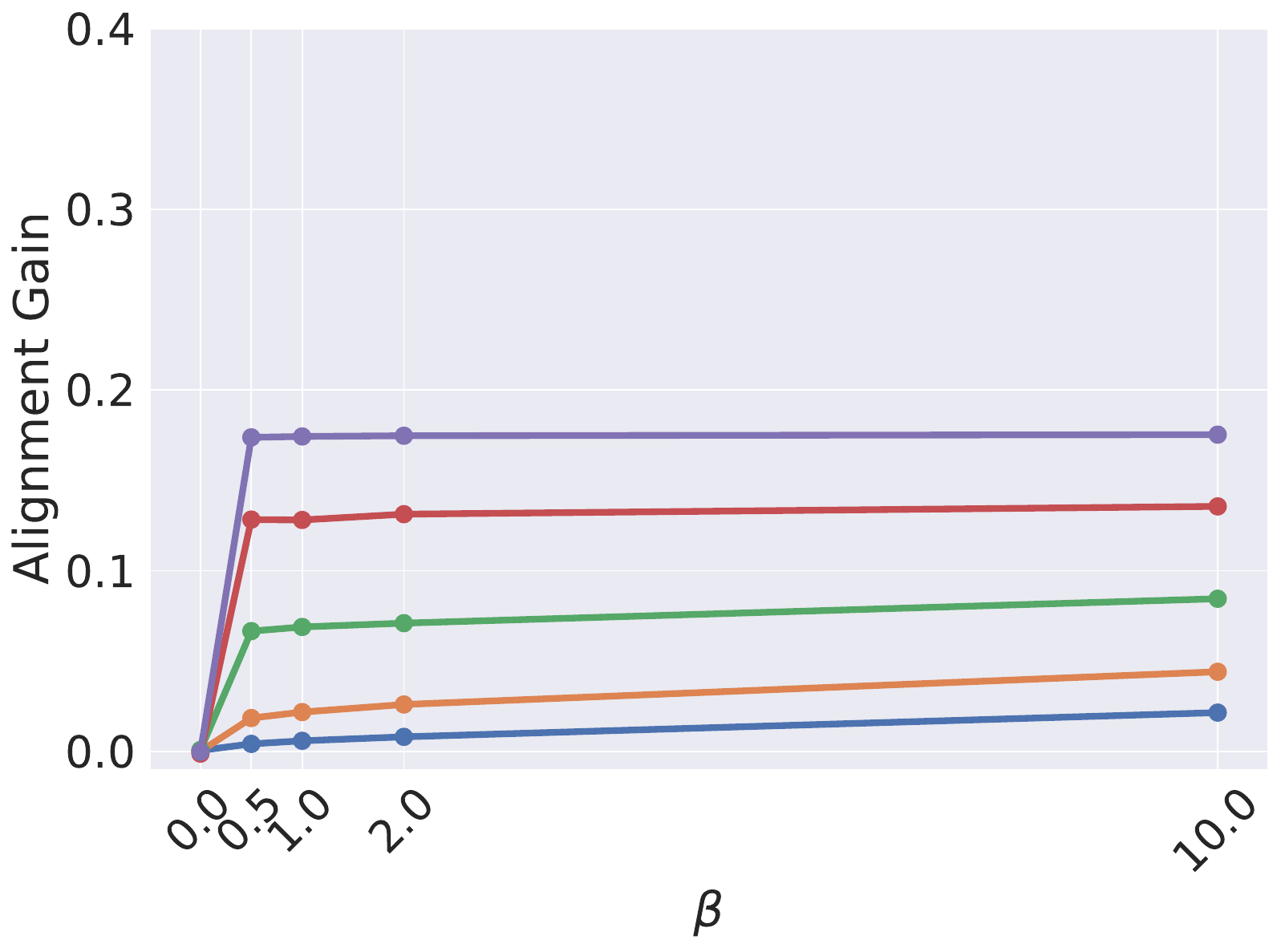}
    \caption{VGG-19, CIFAR-100, WaNet}
  \end{subfigure}

  \medskip

  \begin{subfigure}{0.32\textwidth}
    \centering
    \includegraphics[width=\linewidth]{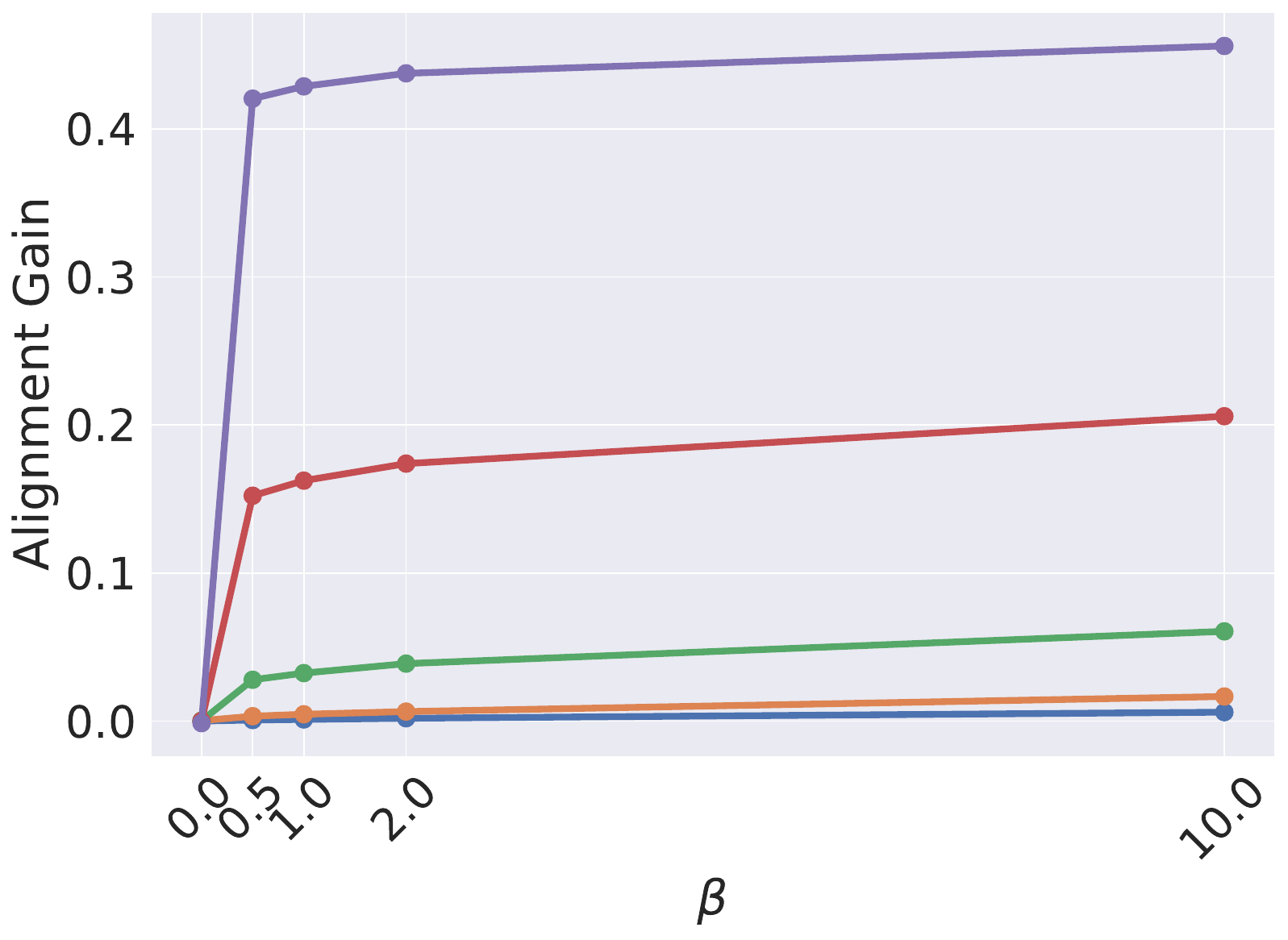}
    \caption{VGG-19, TinyImageNet, BadNets}
  \end{subfigure}
  \hfill
  \begin{subfigure}{0.32\textwidth}
    \centering
    \includegraphics[width=\linewidth]{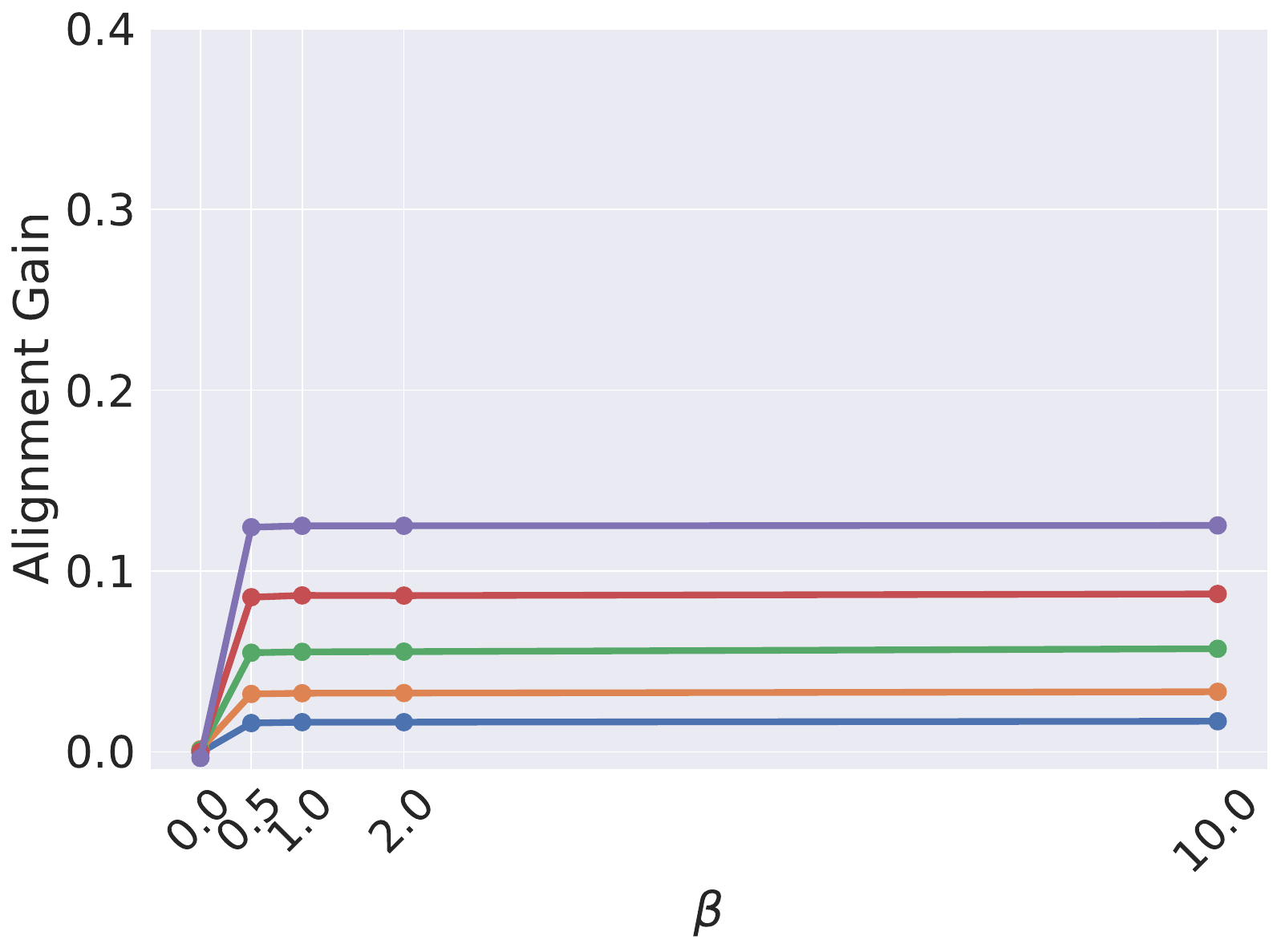}
    \caption{VGG-19, TinyImageNet, Blend}
  \end{subfigure}
  \hfill
  \begin{subfigure}{0.32\textwidth}
    \centering
    \includegraphics[width=\linewidth]{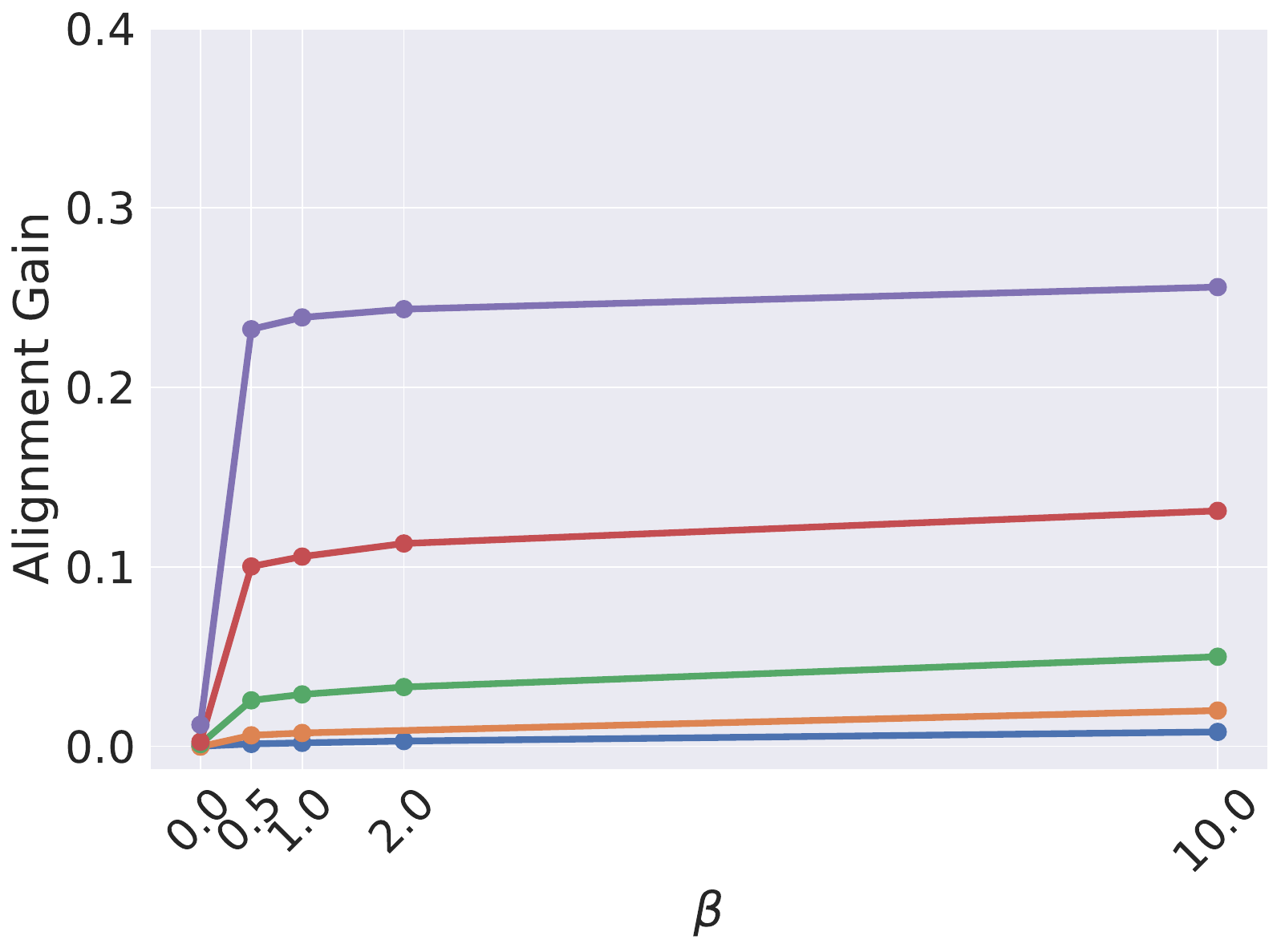}
    \caption{VGG-19, TinyImageNet, WaNet}
  \end{subfigure}

  \caption{Trade-off between alignment and performance for VGG-19 across datasets and attacks. Blue = 2/255, orange = 4/255, green = 8/255, pink = 16/266, and purple = 32/255.}
  \label{fig:vgg19_alignment_tradeoff}
\end{figure*}

Figures~\ref{fig:resnet18_alignment_tradeoff} and~\ref{fig:vgg19_alignment_tradeoff} show the relationship between feature guidance weight $\beta$, alignment with the backdoor direction $d_\ell$, and attack success rate across ResNet-18 and VGG-19. Each subplot presents results for a specific (dataset, attack) combination, with different colors representing perturbation budgets $\varepsilon \in \{2/255, 4/255, 8/255, 16/255, 32/255\}$.

FGA consistently achieves higher alignment than T-PGD across all settings. The alignment gain increases with $\beta$, reaching maximum separation at $\beta=10$. This validates that the feature guidance term successfully constrains the optimization to follow the backdoor mechanism. Larger perturbation budgets amplify the alignment effect due to larger optimization freedom. This suggests that when sufficient perturbation budget is available, FGA can navigate more directly toward the backdoor space than T-PGD. The fact that both T-PGD and FGA achieve comparable ASR and similar alignment indicates that the backdoor region $R_t$ is sufficiently large that multiple optimization strategies converge to it.

\subsection{Perceptual Similarity}
\label{sec:perceptual_sim}

To evaluate the stealthiness of both original and alternative triggers, we measure perceptual similarity using two standard metrics: LPIPS~\cite{zhang2018unreasonable} and SSIM~\cite{wang2004image}.
For LPIPS, we follow common practice and use AlexNet as the feature extractor.
SSIM ranges from $-1$ to $1$, where $1$ indicates perfect similarity and $-1$ indicates completely different images.
LPIPS ranges from $0$ upward, where $0$ denotes perfect similarity. Values exceeding $0.5$ are generally considered perceptible to humans~\cite{ghazanfari2023r}.

We report two sets of measurements: \emph{original trigger similarity}, comparing clean images against poisoned images using the original trigger pattern, and \emph{alternative trigger similarity}, comparing clean images against poisoned images using our FGA-crafted alternative triggers. Experiments are conducted on CIFAR-10, CIFAR-100, and TinyImageNet using ResNet-18, across four backdoor attacks: BadNets, WaNet, Blend, and Input-Aware. For the alternative triggers, we report results for three perturbation budgets $\varepsilon \in \{8/255,\, 16/255,\, 32/255\}$.

Results are summarized in Table~\ref{tab:perceptual_similarity}. Original triggers for BadNets, WaNet, and Blend achieve high SSIM scores ($> 0.49$) and low LPIPS values (${<}\,0.11$ on CIFAR; ${<}\,0.11$ on TinyImageNet).
Input-Aware triggers are a notable exception, with very low SSIM ($<\,0.03$) and high LPIPS (up to $0.59$ on TinyImageNet), indicating they introduce perceptible distortions. In contrast, our alternative triggers remain consistently stealthy across all settings, with LPIPS values below $0.29$ on all datasets and all perturbation budgets, well within the imperceptibility threshold.

\begin{table}[htb]
  \centering
  \caption{Perceptual similarity of original triggers (clean vs.\ poisoned) and alternative triggers (clean vs.\ adversarially-crafted) across datasets, attack types, and perturbation budgets $\varepsilon$. Each adversarial cell reports LPIPS\,/\,SSIM.}
  \label{tab:perceptual_similarity}
  \resizebox{\linewidth}{!}{%
  \begin{tabular}{llccccc}
    \toprule
    \multirow{2}{*}{\textbf{Dataset}}
      & \multirow{2}{*}{\textbf{Trigger}}
      & \multicolumn{2}{c}{\textbf{Original Trigger}}
      & \multicolumn{3}{c}{\textbf{Alternative Trigger (LPIPS / SSIM)}} \\
    \cmidrule(lr){3-4}\cmidrule(lr){5-7}
      & & LPIPS$\downarrow$ & SSIM$\uparrow$ & $\varepsilon=8/255$ & $\varepsilon=16/255$ & $\varepsilon=32/255$ \\
    \midrule
  \multirow{4}{*}{CIFAR-10} & BadNets & 0.018 & 0.942 & 0.127 / 0.252 & 0.127 / 0.236 & 0.127 / 0.219 \\
   & WaNet & 0.007 & 0.943 & 0.127 / 0.252 & 0.127 / 0.234 & 0.126 / 0.216 \\
   & Blend & 0.019 & 0.495 & 0.127 / 0.252 & 0.127 / 0.233 & 0.128 / 0.213 \\
   & Input-Aware & 0.236 & 0.014 & 0.128 / 0.249 & 0.131 / 0.227 & 0.137 / 0.194 \\
  \midrule
  \multirow{4}{*}{CIFAR-100} & BadNets & 0.020 & 0.942 & 0.129 / 0.274 & 0.130 / 0.259 & 0.130 / 0.241 \\
   & WaNet & 0.006 & 0.950 & 0.129 / 0.273 & 0.129 / 0.258 & 0.128 / 0.239 \\
   & Blend & 0.020 & 0.534 & 0.129 / 0.272 & 0.130 / 0.256 & 0.131 / 0.235 \\
   & Input-Aware & 0.264 & 0.010 & 0.129 / 0.269 & 0.131 / 0.248 & 0.135 / 0.211 \\
  \midrule
  \multirow{4}{*}{TinyImageNet} & BadNets & 0.107 & 0.937 & 0.278 / 0.180 & 0.279 / 0.175 & 0.281 / 0.169 \\
   & WaNet & 0.026 & 0.935 & 0.279 / 0.180 & 0.280 / 0.174 & 0.282 / 0.168 \\
   & Blend & 0.015 & 0.751 & 0.278 / 0.180 & 0.278 / 0.174 & 0.280 / 0.166 \\
   & Input-Aware & 0.591 & 0.029 & 0.278 / 0.173 & 0.280 / 0.166 & 0.287 / 0.156 \\
    \bottomrule
  \end{tabular}}
\end{table}

\end{document}